\documentclass[10pt,journal,compsoc]{IEEEtran}
\ifCLASSOPTIONcompsoc
\usepackage[nocompress]{cite}
\else
\usepackage{cite}
\fi

\ifCLASSINFOpdf
\else
\fi

\usepackage{url}            
\usepackage{booktabs}       
\usepackage{amsfonts}       
\usepackage{nicefrac}       
\usepackage{microtype}      
\usepackage{xcolor}         

\usepackage{graphicx}
\usepackage{subfigure}
\usepackage{amsmath}
\usepackage{amssymb}
\usepackage{mathtools}
\usepackage{amsthm}
\usepackage{bm}
\usepackage{wrapfig}
\usepackage{multirow} 
\usepackage{makecell}

\theoremstyle{plain}
\newtheorem{theorem}{Theorem}
\newtheorem{proposition}{Proposition}
\newtheorem{lemma}{Lemma}
\newtheorem{corollary}{Corollary}
\theoremstyle{definition}
\newtheorem{definition}{Definition}
\newtheorem{assumption}{Assumption}
\theoremstyle{remark}
\newtheorem{remark}{Remark}

\begin{document}
	
	\title{Towards Understanding the Generalizability of Delayed Stochastic Gradient Descent}
	
	
	\author{Xiaoge~Deng,
		Li~Shen,
		Shengwei~Li,
		Tao~Sun,
		Dongsheng~Li
		and~Dacheng Tao, ~\IEEEmembership{Fellow,~IEEE}
		\IEEEcompsocitemizethanks{
			\IEEEcompsocthanksitem X. Deng is with the Intelligent Game and Decision Lab, Beijing, China. S. Li, T. Sun and D. Li are with the College of Computer Science and Technology, National University of Defense Technology, Hunan, China. E-mail: dengxg@nudt.edu.cn, lucasleesw9@gmail.com, suntao.saltfish@outlook.com, dsli@nudt.edu.cn.
			\IEEEcompsocthanksitem L. Shen is with the Sun Yat-sen University, Guangzhou, China. E-mail: mathshenli@gmail.com. D. Tao is with the JD Explore Academy. E-mail: dacheng.tao@gmail.com.
			
			\IEEEcompsocthanksitem Corresponding author: Tao Sun.}
	}
	
	\markboth{Journal of \LaTeX\ Class Files,~Vol.~14, No.~8, August~2015}%
	{Shell \MakeLowercase{\textit{et al.}}: Bare Demo of IEEEtran.cls for Computer Society Journals}

	\IEEEtitleabstractindextext{%
		\begin{abstract}
			Stochastic gradient descent (SGD) performed in an asynchronous manner plays a crucial role in training large-scale machine learning models. However, the generalization performance of asynchronous delayed SGD, which is an essential metric for assessing machine learning algorithms, has rarely been explored. Existing generalization error bounds are rather pessimistic and cannot reveal the correlation between asynchronous delays and generalization. In this paper, we investigate sharper generalization error bound for SGD with asynchronous delay $\tau$. Leveraging the generating function analysis tool, we first establish the average stability of the delayed gradient algorithm. Based on this algorithmic stability, we provide upper bounds on the generalization error of $\widetilde{\mathcal{O}}(\frac{T-\tau}{n\tau})$ and $\widetilde{\mathcal{O}}(\frac{1}{n})$ for quadratic convex and strongly convex problems, respectively, where $T$ refers to the iteration number and $n$ is the amount of training data. Our theoretical results indicate that asynchronous delays reduce the generalization error of the delayed SGD algorithm. Analogous analysis can be generalized to the random delay setting, and the experimental results validate our theoretical findings.
		\end{abstract}
		
		}

	\maketitle

	\IEEEdisplaynontitleabstractindextext
	\IEEEpeerreviewmaketitle

	\IEEEraisesectionheading{\section{Introduction}\label{sec:introduction}}
	
	\IEEEPARstart{F}{irst}-order gradient-based optimization methods, such as stochastic gradient descent (SGD), are the mainstay of supervised machine learning (ML) training today \cite{robbins1951stochastic, bottou2018optimization}. The plain SGD starts from an initial point $\mathbf{w}_0$ and trains iteratively on the training datasets by $\mathbf{w}_{t+1}=\mathbf{w}_t-\eta \mathbf{g}_{t}$, where $\mathbf{w}_t$ denotes the current iteration, $\eta$ is the learning rate, and $\mathbf{g}_{t}$ represents the gradient evaluated at $\mathbf{w}_t$. In practice, SGD not only learns convergent models efficiently on training datasets, but surprisingly, the learned solutions perform well on unknown testing data, exhibiting good generalization performance \cite{zhang2017understanding, du2018gradient}. Generalizability is a focal topic in the ML community, and there has been considerable effort to investigate the generalization performance of gradient-based optimization methods from the algorithmic stability perspective \cite{hardt2016train, kuzborskij2018data, lei2020fine, bassily2020stability, zhang2022stability, zhou2022understanding}.
	
	However, in modern ML applications, the scale of samples in the training datasets and the number of parameters in the deep neural network models are so large that it is imperative to implement gradient methods in a distributed manner \cite{deng2009imagenet, dean2012large, brown2020language}. Consider a distributed parameter server training system with $M$ workers and the distributed SGD performed as $\mathbf{w}_{t+1}=\mathbf{w}_{t}-\eta \sum_{m=1}^{M}\mathbf{g}_{t}^{m}/M$, where $\mathbf{g}_{t}^{m}$ represents the gradient evaluated by the $m$-th worker at the $t$-th iteration \cite{zinkevich2010parallelized, li2014scaling}. This straightforward distributed implementation synchronizes the gradients from all workers at each iteration, which facilitates algorithmic convergence but imposes a significant synchronization overhead on the distributed training system \cite{assran2020advances}.

	An alternative approach to tackling this synchronization issue is employing asynchronous training, which eliminates the synchronization barrier and updates the model whenever the server receives gradient information without waiting for any other workers \cite{nedic2001distributed, agarwal2011distributed, lian2015asynchronous}. It is essential to note that when the server receives the gradient data from a particular worker, the global model parameter has undergone several asynchronous updates, making asynchronous training a delayed gradient update. Hence the delayed SGD performs as $\mathbf{w}_{t+1}=\mathbf{w}_t-\eta \mathbf{g}_{t-\tau}$, where $\tau$ denotes the asynchronous delay and $\mathbf{g}_{t-\tau}$ is the gradient evaluated at $\mathbf{w}_{t-\tau}$ (as shown in Figure \ref{delayed_sgd}). Despite the existence of delayed updates, extensive theoretical work has been developed to guarantee the convergence of delayed gradient methods \cite{zhou2018distributed, arjevani2020tight, stich2020error, cohen2021asynchronous, mishchenko2022asynchronous}.

\begin{figure}[!t]
		\centering
		\includegraphics[width=2.68in]{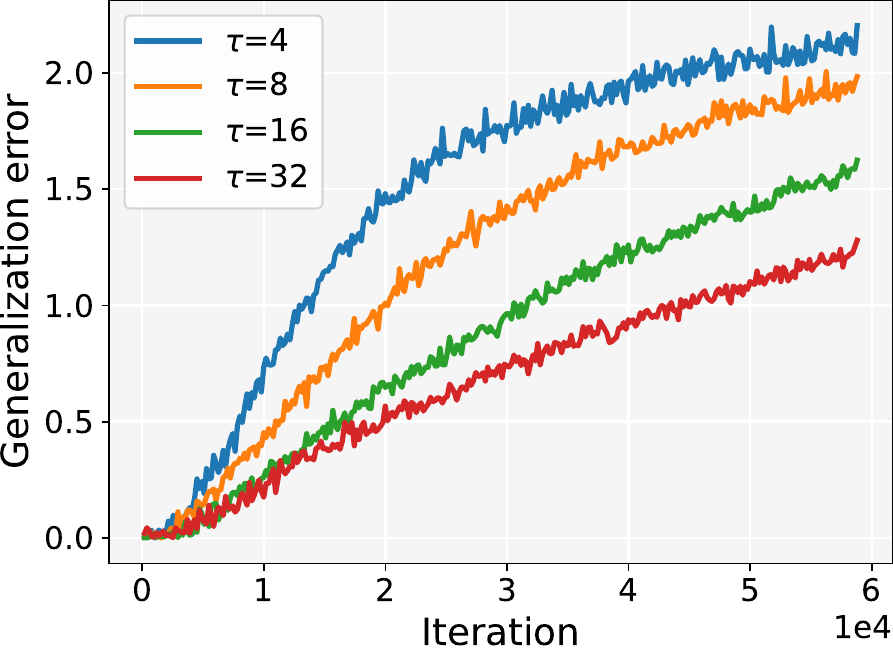}
		\caption{Generalization error of the ResNet-18 model trained with delayed SGD for classifying the CIFAR-100 data set. The experiment varied only the asynchronous delays and fixed other parameters, where the learning rate $\eta = 0.1$.}
		\label{fig:gen_resnet}
	\end{figure}

	While the convergence properties of delayed gradient methods have been widely studied, their generalization performance has yet to be explored, and generalization is the ultimate goal of the ML community. \cite{regatti2019distributed} analyzed the generalization performance of the delayed SGD algorithm in the non-convex case with the uniform stability framework, and gave an upper bound on the generalization error of $\mathcal{O}(T^{\tau}/n\tau)$, where $T$ is the iteration number and $n$ denotes the number of training samples. Although Regatti et al. claim that ``distributed SGD generalizes well under asynchrony" in \cite{regatti2019distributed}, their theoretical upper bound is rather pessimistic. Moreover, empirical evidence suggests that increasing the asynchronous delay reduces the generalization error at an appropriate learning rate, as illustrated in Figure \ref{fig:gen_resnet}, which cannot be explained by the results of \cite{regatti2019distributed}. Therefore, there is an imperative demand for a theoretical study into the inherent connection between asynchronous delay and generalization error for gradient descent algorithms.
	
	This paper is devoted to understanding the effect of asynchronous delays on the generalization error of the delayed stochastic gradient descent algorithm. We focus on a convex quadratic optimization problem, which is a fundamental and critical topic in the ML community. Empirical evidence demonstrates that the local regions around the minimum of non-convex deep neural networks are usually convex \cite{li2018visualizing}, and utilizing local quadratic approximations around the minimum is a fruitful way to study the behavior of deep neural networks \cite{jacot2018neural, ma2018power, he2019control, zou2021benign}. In this study, we investigate the data-dependent average stability of delayed SGD to estimate the generalization error of the algorithm. Recursive sequences related to the algorithmic stability are established for the quadratic optimization problem. It is worth noting that the established recursive property takes the form of an equational relation. Consequently, we utilize these sequences as coefficients to construct the corresponding generating functions for analysis, yielding tighter generalization error bounds. Our contributions are summarized as follows.
	
	\begin{itemize}
		\item We provide sharper upper bounds on the generalization error of the delayed stochastic gradient descent algorithm. By studying the average stability of the delayed gradient algorithm using generating functions, we derive generalization results. Our findings indicate that asynchronous delays can reduce the generalization error at appropriate learning rates.
		
		\item For the convex quadratic problem, we present the upper generalization error bound of $\widetilde{\mathcal{O}}(\frac{T-\tau}{n\tau})$ for delayed SGD with delay $\tau$. When the quadratic function is strongly convex, we derive the generalization error bound $\widetilde{\mathcal{O}}(\frac{1}{n})$ that is independent of iteration numbers $T$.\footnote{Notation $\mathcal{O}$ represents hiding constants, and tilde $\widetilde{\mathcal{O}}$ hides both constants and polylogarithmic factors of $\tau$.}
		
		\item We extend the analysis to time-varying random delay settings and yield analogous generalization error bounds, namely $\widetilde{\mathcal{O}}(\frac{T-\overline{\tau}}{n\overline{\tau}})$ and $\widetilde{\mathcal{O}}(\frac{1}{n})$ in the convex and strongly convex cases, respectively, where $\overline{\tau}$ is the maximum delay. Notably, all our bounds explicitly demonstrate that increasing the amount of training data $n$ improves the generalization performance.
		
		\item We conducted exhaustive experiments showing that asynchronous delays reduce the generalization error of delayed SGD, which is consistent with our theoretical findings.\footnote{Code is available at \url{github.com/xiaogdeng/Gen-DelayedSGD}.}
	\end{itemize}

	\section{Related work}
	\label{sec:related}
	\noindent \textbf{Delayed gradient methods.$~$} Asynchronous training can be traced back to \cite{tsitsiklis1986distributed}, enabling training with delayed gradients and allowing for tolerance to struggling problems. A widely recognized method is asynchronous training in shared memory, such as Hogwild! \cite{recht2011hogwild}, but it is unsuitable for modern ML applications and is not our focus. This paper examines the situation of asynchronous training with delayed gradients in a distributed-memory architecture, which is closely related to \cite{nedic2001distributed, agarwal2011distributed}. Extensive researches have been conducted to analyze the convergence of such delayed gradient algorithms \cite{lian2015asynchronous, mania2015perturbed, sun2017asynchronous, zhou2018distributed}, and to provide the optimal rates for various circumstances \cite{arjevani2020tight, stich2020error, aviv2021asynchronous}. Additionally, recent studies have demonstrated that the asynchronous gradient method is robust to arbitrary delays \cite{cohen2021asynchronous, mishchenko2022asynchronous}. There are also some efforts to enhance the performance of asynchronous algorithms with gradient compensation \cite{zheng2017asynchronous} and delay-adaptive learning rate \cite{sra2016adadelay, zhang2016staleness, ren2020delay, backstrom2022asap}. Moreover, delayed gradient algorithms are particularly popular in reinforcement learning \cite{mnih2016asynchronous}, federated learning \cite{koloskova2022sharper}, and online learning \cite{hsieh2022multi} fields.
	
	\noindent \textbf{Stability and generalization.$~$} Algorithm stability is based on sensitivity analysis, which measures how the learning algorithm reacts to perturbations in the training datasets \cite{rogers1978finite, devroye1979distribution1, devroye1979distribution2}. The foundational work \cite{bousquet2002stability} defined three algorithmic stability notations and established its relation to generalization performance. This framework was subsequently enhanced to cover randomized learning algorithms \cite{elisseeff2005stability}. Stability is also a key necessary and sufficient condition for learnability \cite{mukherjee2006learning, shalev2010learnability}. The seminal work \cite{hardt2016train} studied the generalization error of SGD using uniform stability tools and motivated a series of extended researches \cite{charles2018stability, bassily2020stability, zhang2022stability}. However, uniform stability usually yields worst-case generalization error bounds, hence \cite{shalev2010learnability, kuzborskij2018data, lei2020fine} explored data-dependent on-average stability and  obtained improved results. Besides in the expectation sense, algorithmic stability is also used to study the high probability bounds of generalization \cite{feldman2018generalization, feldman2019high, bousquet2020sharper}. Recently, \cite{chandramoorthy2022on} introduced statistical algorithmic stability and studied the generalizability of non-convergent algorithms, and \cite{richards2021stability, lei2022stability} explored the generalization performance of gradient methods in overparameterized neural networks with stability tools. Stability-based generalization analysis has also been applied to Langevin dynamics \cite{mou2018generalization, banerjee2022stability}, pairwise learning \cite{lei2020sharper, yang2021simple}, meta learning \cite{farid2021generalization, guan2022finegrained} and minimax problems \cite{farnia2021train, lei2021stability, xing2021algorithmic, NEURIPS2022_f9b8853e, xiao2022stability}. In the distributed settings, \cite{wu2019stability} defined uniform distributed stability and investigated the generalization performance of divide-and-conquer distributed algorithms. Recently, \cite{sun2021stability, deng2023stability} and \cite{zhu2022topology} studied the generalization performance of distributed decentralized SGD algorithms based on uniform stability and on-average stability, respectively.
	
	Under the uniform stability framework, Regatti et al. \cite{regatti2019distributed} presented a conservative generalization error bound $\mathcal{O}(L^{2}T^{\tau}/n \tau)$ for delayed SGD in non-convex settings, where $L$ denotes the gradient bound. In this paper, we aim at studying quadratic convex problems by using average stability. Without relying on the bounded gradient assumption, we demonstrate that asynchronous delays make the stochastic gradient descent algorithm more stable and hence reduce the generalization error.

	\section{Preliminaries}
	\label{sec:problem}
	This section outlines the problem setup and describes the delayed gradient methods, followed by an introduction of priori knowledge about stability and generalization. Throughout this paper, we will use the following notation.
	
	\noindent \textbf{Notation.} Bold capital and lowercase letters represent the matrices and column vectors, respectively. Additionally, $(\cdot)^{\top}$ denotes the transpose of the corresponding matrix or vector. For a vector $\mathbf{x}\in \mathbb{R}^{d}$, $\|\mathbf{x}\|$ represents its $\ell_{2}$-norm. The integer set $\{1, 2, \ldots, d\}$ is represented by $[d]$. Lastly, $\mathbb{E}[\cdot]$ denotes the expectation of a random variable with respect to the underlying probability space. 
	
	\subsection{Problem formulation}
	In this paper, we consider the general supervised learning problem, where $\mathcal{X}\in\mathbb{R}^{d}$ and $\mathcal{Y}\in\mathbb{R}$ are the input and output spaces, respectively. Let $\mathcal{S}=\{\mathbf{z}_{1}=(\mathbf{x}_{1}, y_1), \ldots, \mathbf{z}_{n}=(\mathbf{x}_{n}, y_n)\}$ be a training set of $n$ examples in $\mathcal{Z}=\mathcal{X}\times\mathcal{Y}$, drawn independent and identically distributed (i.i.d.) from an unknown distribution $\mathcal{D}$. Denote the loss of model $\mathbf{w}$ on sample $\mathbf{z}$ as $f(\mathbf{w}; \mathbf{z})$. Specifically, we focus on the convex quadratic loss function $f(\mathbf{w}; \mathbf{z}_{i})=f(\mathbf{w};\mathbf{x}_{i}, y_i)=\frac{1}{2}(\mathbf{x}_{i}^{\top}\mathbf{w}-y_i)^{2}$, where $\mathbf{w}\in\mathbb{R}^{d}$ is the learning model. The empirical risk minimization (ERM) problem can be formulated as
	\begin{equation}
		\label{erm}
		\min_{\mathbf{w}\in\mathbb{R}^{d}} ~ F_{\mathcal{S}}(\mathbf{w})=\frac{1}{n}\sum_{i=1}^{n}f(\mathbf{w};\mathbf{z}_{i})=\frac{1}{2}\mathbf{w}^{\top}\mathbf{A} \mathbf{w}+\mathbf{b}^{\top}\mathbf{w}+c,
	\end{equation}
	where $\mathbf{A}=\frac{1}{n}\sum_{i=1}^{n}\mathbf{x}_{i}\mathbf{x}_{i}^{\top}\in\mathbb{R}^{d\times d}$ is a positive semi-definite matrix with eigenvalues $\{a_{1}, \ldots, a_{d}\}$. $\mathbf{b}=\frac{1}{n}\sum_{i=1}^{n}-y_i\mathbf{x}_{i}\in\mathbb{R}^{d}$, and $c=\frac{1}{2n}\sum_{i=1}^{n}y_i^2\in\mathbb{R}$ is a scalar.
	
	The main objective of ML algorithms is to minimize the population risk $F(\mathbf{w})=\mathbb{E}_{\mathbf{z}\sim\mathcal{D}}[f(\mathbf{w};\mathbf{z})]$, which denotes the expected risk of the model. Unfortunately, we cannot directly compute $F(\mathbf{w})$ since the distribution $\mathcal{D}$ is unknown. In practice, we instead solve the approximated empirical risk $F_{\mathcal{S}}(\mathbf{w})$. This paper aims to investigate the disparity between the population risk and the empirical risk, referred to as the generalization error. Specifically, for a given algorithm $\mathcal{A}$, $\mathbf{w}=\mathcal{A}(\mathcal{S})$ denote the output model obtained by minimizing the empirical risk on the training data set $\mathcal{S}$ with $\mathcal{A}$. The \textit{generalization error} $\epsilon_{\text{gen}}$ is defined as the expected difference between the population risk and the empirical risk, where the expectation is taken over the randomness of algorithm and training samples.
	\begin{equation}
		\nonumber
		\epsilon_{\text{gen}}=\mathbb{E}_{\mathcal{S}, \mathcal{A}}\left[F(\mathcal{A}(\mathcal{S}))-F_{\mathcal{S}}(\mathcal{A}(\mathcal{S}))\right].
	\end{equation}
	
	\subsection{Delayed gradient methods}
	\begin{figure}[!t]
		\centering
		\includegraphics[width=0.45\textwidth]{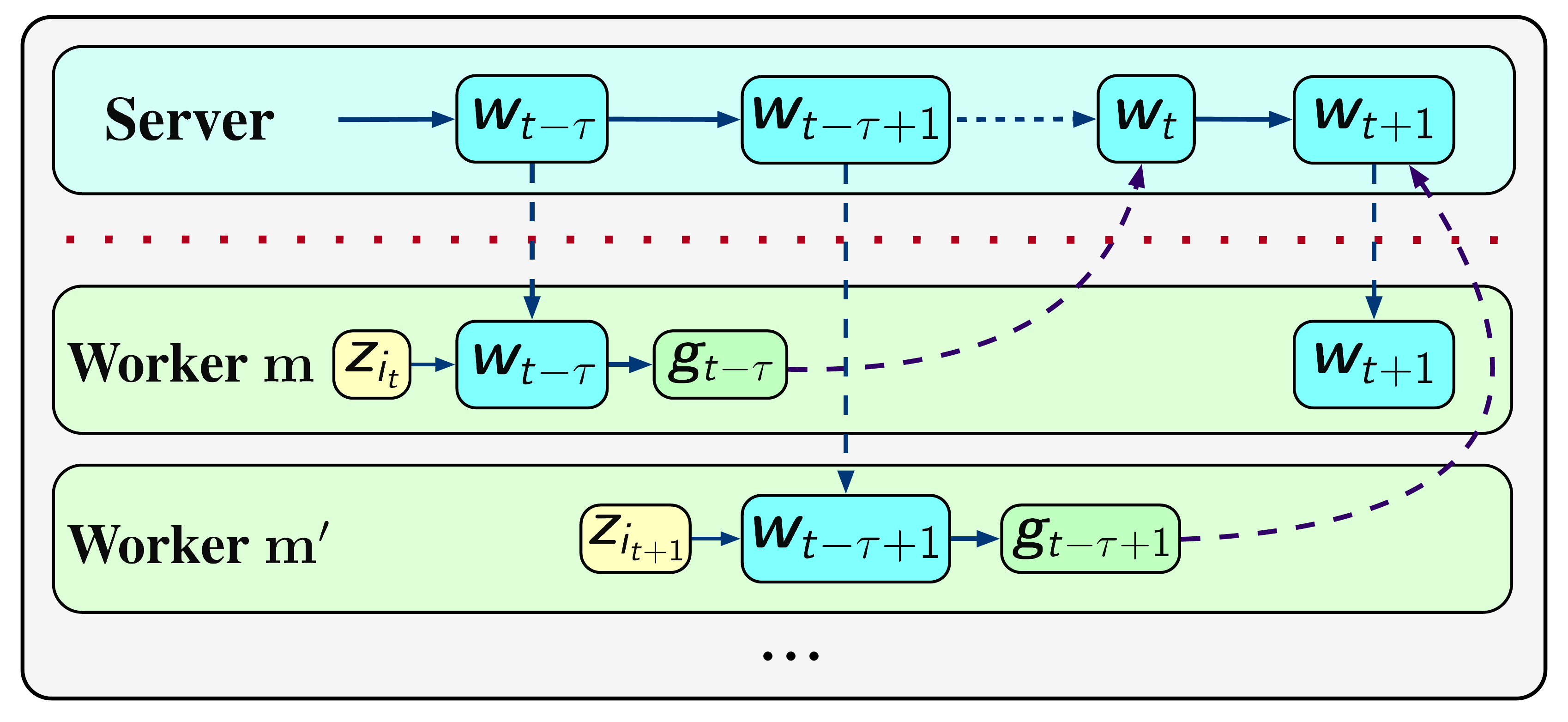}
		\caption{Schematic of the delayed gradient algorithms. While worker $m$ is computing and uploading the gradient, the server has performed $\tau$ asynchronous model updates.}
		\label{delayed_sgd}
	\end{figure}
	To find a solution for the ERM problem \eqref{erm}, delayed gradient methods initialize the model parameters to $\mathbf{w}_{0}=\mathbf{w}_{1}=\cdots=\mathbf{w}_{\tau}$ and then iteratively update them with gradient descent. Figure \ref{delayed_sgd} illustrates that the model $\mathbf{w}_{t}$ is updated with the stale gradient $\mathbf{g}_{t-\tau}$ at the $t$-th iteration, implying an iterative format of $\mathbf{w}_{t+1}=\mathbf{w}_t-\eta \mathbf{g}_{t-\tau}$ where $t>\tau$. In SGD, the delayed gradient $\mathbf{g}_{t-\tau}$ is evaluated on a randomly selected data point $\mathbf{z}_{i_t}$, and also constitutes a noisy gradient oracle of the empirical risk $F_\mathcal{S}$, i.e.,
	\begin{equation}
		\label{grad_format}
		\mathbf{g}_{t-\tau}=\nabla f(\mathbf{w}_{t-\tau}; \mathbf{z}_{i_t})=\nabla F_{\mathcal{S}}(\mathbf{w}_{t-\tau})+\bm{\xi}_{t},
	\end{equation}
	where $\bm{\xi}_{t}$ represents the associated stochastic noise. \textit{Delayed SGD} updates the model parameters as
	\begin{equation}
		\label{def:sgd}
		\begin{split}
			\mathbf{w}_{t+1}&=\mathbf{w}_{t}-\eta \nabla f(\mathbf{w}_{t-\tau}; \mathbf{z}_{i_t})\\
			&=\mathbf{w}_{t}-\eta\left(\nabla F_{\mathcal{S}}(\mathbf{w}_{t-\tau})+\bm{\xi}_{t}\right).
		\end{split}
	\end{equation}
	
	\subsection{Stability and generalization}
	Algorithm stability is a powerful tool for studying generalization. It estimates the generalization error of an algorithm by assessing the effect of changing a single sample in the training set on the output. Denote by $\mathcal{S}'=\{\mathbf{z}_{1}', \ldots, \mathbf{z}_{n}'\}$ the set of random samples drawn i.i.d. from distribution $\mathcal{D}$ and independent of $\mathcal{S}$. Let $\mathcal{S}^{(i)}=\{\mathbf{z}_{1}, \ldots, \mathbf{z}_{i-1}, \mathbf{z}_{i}', \mathbf{z}_{i+1}, \ldots, \mathbf{z}_{n}\}$ be the replica of the sample $\mathcal{S}$ except in the $i$-th example, where we substitute $\mathbf{z}_{i}$ with $\mathbf{z}_{i}'$. In this paper, we employ the following \textit{average stability} framework \cite{shalev2010learnability}.
	\begin{definition}[average stability]
		\label{def:1}
		A stochastic learning algorithm $\mathcal{A}$ is $\epsilon_{\text{stab}}$-average stable if for any datasets $\mathcal{S}$, $\mathcal{S}^{(i)}$ which differ in one example, we have
		\begin{equation}
			\nonumber
			\left|\frac{1}{n}\sum_{i=1}^{n}\mathbb{E}_{\mathcal{S}, \mathcal{S}', \mathcal{A}}\left[f(\mathcal{A}(\mathcal{S});\mathbf{z}_{i}')-f(\mathcal{A}(\mathcal{S}^{(i)});\mathbf{z}_{i}')\right]\right|\leq \epsilon_{\text{stab}}.
		\end{equation}
	\end{definition}
	The following lemma establishes the connection between average stability and generalization error \cite[Lemma 11]{shalev2010learnability}.
	\begin{lemma}
		\label{lem:1}
		Let algorithm $\mathcal{A}$ be $\epsilon_{\text{stab}}$-average stable. Then the generalization error satisfies
		\begin{equation}
			\nonumber
			\left|\mathbb{E}_{\mathcal{S}, \mathcal{A}}\left[F(\mathcal{A}(\mathcal{S}))-F_{\mathcal{S}}(\mathcal{A}(\mathcal{S}))\right]\right| \leq \epsilon_{\text{stab}}.
		\end{equation}
	\end{lemma}
	\begin{remark}
		\label{rmk:0}
		Since for any $i$, $\mathbf{z}_{i}$ and $\mathbf{z}_{i}'$ are both drawn i.i.d. from distribution $\mathcal{D}$, then $\mathcal{A}(\mathcal{S}^{(i)})$ is independent of $\mathbf{z}_{i}$, and the average stability can be formulated as
		\begin{equation}
			\label{new_stab}
			\left|\frac{1}{n}\sum_{i=1}^{n}\mathbb{E}_{\mathcal{S}, \mathcal{S}', \mathcal{A}}\left[f(\mathcal{A}(\mathcal{S}^{(i)});\mathbf{z}_{i})-f(\mathcal{A}(\mathcal{S});\mathbf{z}_{i})\right]\right|\leq \epsilon_{\text{stab}}.
		\end{equation}
		This formulation is equivalent to Definition \ref{def:1} and shares the property outlined in Lemma \ref{lem:1}. We have provided a detailed analytical proof in the Supplementary material.
	\end{remark}
	Therefore, our goal turns to study the average stability of the algorithm, which is sufficient to bound the generalization error. Before exploring the average stability and generalization error of the delayed stochastic gradient descent algorithm, we make the following regularity assumptions.
	\begin{assumption}[smoothness]
		\label{amp:1}
		The loss function $F_{\mathcal{S}}$ is $\mu$-smooth, i.e., there exists a constant $\mu\in\mathbb{R}$ such that for any $\mathbf{w}, \mathbf{v}\in\mathbb{R}^{d}$,
		\begin{equation}
			\nonumber
			\|\nabla F_{\mathcal{S}}(\mathbf{w})-\nabla F_{\mathcal{S}}(\mathbf{v})\|\leq\mu\|\mathbf{w}-\mathbf{v}\|.
		\end{equation}
	\end{assumption}
	
	\begin{assumption}[bounded space]
		\label{amp:2}
		The mean of the sample points in the training data set is bounded, i.e., $\|\mathbf{b}\|\leq r$, where $r\in\mathbb{R}$ is a constant, and $\mathbf{b}=\frac{1}{n}\sum_{i=1}^{n}-y_i\mathbf{x}_{i}\in\mathbb{R}^{d}$.
	\end{assumption}
	
	\begin{assumption}[bounded noise]
		\label{amp:3}
		The stochastic noise $\bm{\xi}_{t}$ has a bounded second-order moment with respect to the training set $\mathcal{S}$ and stochastic algorithm $\mathcal{A}$, i.e., there exists a constant $\sigma>0$ such that $\mathbb{E}_{\mathcal{S}, \mathcal{A}}[\|\bm{\xi}_{t}\|^{2}]\leq \sigma^{2}$.
	\end{assumption}
	
	Assumption \ref{amp:1} is to bound the eigenvalues of the matrix $\mathbf{A}$, i.e., $\max_{j\in[d]} a_j < \mu$, which is natural in quadratic problems. Assumption \ref{amp:2} typically holds in supervised machine learning applications. In stochastic algorithms, it is common to assume that the random noise is bounded, independent, and has zero mean \cite{arjevani2020tight, stich2020error, aviv2021asynchronous}. However, in this paper, we only require the noise to be bounded (Assumption \ref{amp:3}).

	\section{Average stability via generating function derivations}
	\label{sec:stab}
	This section focuses on analyzing the average stability of the delayed stochastic gradient descent algorithm with fixed delay $\tau$, and the derivation is generalized to random delays in Section \ref{sec:random_delay}. To measure the algorithmic stability, we run the delayed SGD algorithm on two datasets $\mathcal{S}$ and $\mathcal{S}^{(i)}$ from the same starting point, and obtain the models $\mathbf{w}_{t}$ and $\mathbf{w}_{t}'$ after $t$ iterations, respectively. According to Definition \eqref{new_stab}, the average stability under the quadratic loss function can be formulated as
	\begin{equation}
		\label{avg_stab}
		\begin{split}
			&\mathbb{E}\left|\frac{1}{n}\sum_{i=1}^{n} \left[f(\mathbf{w}_t';\mathbf{z}_{i})-f(\mathbf{w}_t;\mathbf{z}_{i})\right]\right|\\
			&=\mathbb{E}\left|\frac{1}{2}(\mathbf{w}_t'-\mathbf{w}_t)^{\top}\mathbf{A}(\mathbf{w}_t'+\mathbf{w}_t)+\mathbf{b}^{\top}(\mathbf{w}_t'-\mathbf{w}_t)\right|\\
			&\leq\frac{1}{2}\mathbb{E}\left|(\mathbf{w}_t'-\mathbf{w}_t)^{\top}\mathbf{A}(\mathbf{w}_t'+\mathbf{w}_t)\right|+\mathbb{E}\|\mathbf{b}\|\|\mathbf{w}_t'-\mathbf{w}_t\|\\
			&\leq \frac{1}{2}\mathbb{E}\left\|\sqrt{\mathbf{A}}\mathbf{e}_{t}\right\|\left\|\sqrt{\mathbf{A}}\mathbf{s}_{t}\right\|+\mathbb{E}\|\mathbf{b}\|\|\mathbf{e}_{t}\|.
		\end{split}
	\end{equation}
	Here we defined two crucial symbols $\mathbf{e}_{t}=\mathbf{w}_{t}-\mathbf{w}_{t}'$ and $\mathbf{s}_{t}=\mathbf{w}_{t}+\mathbf{w}_{t}'$. It is worth noting that this section investigates the algorithmic stability, hence all the notations $\mathbb{E}$ in this section take expectations with respect to the randomness of $\mathcal{S}, \mathcal{S}'$ and $\mathcal{A}$, consistent with Definition \ref{def:1}. Since the two iterations start from the same model, i.e., $\mathbf{w}_{0}=\mathbf{w}_{0}'$, and combined with the initial setup $\mathbf{w}_{0}=\cdots=\mathbf{w}_{\tau}$ yields
	\begin{equation}
		\label{init}
		\begin{split}
			\mathbf{e}_{0}=\mathbf{e}_{1}=\cdots=\mathbf{e}_{\tau}=\mathbf{0} ~~ \text{and} ~~ \mathbf{s}_{0}=\mathbf{s}_{1}=\cdots=\mathbf{s}_{\tau}.
		\end{split}
	\end{equation}
	
	Recalling that $\mathbf{w}_{t+1}$ is obtained from $\mathbf{w}_{t}$ via delayed gradient updates \eqref{def:sgd}, our goal is to establish recursive properties of the sequences $\{\mathbf{e}_{t}\}_{t}$ and $\{\mathbf{s}_{t}\}_{t}$ based on the iterative behavior of the algorithm. Two cases arise at each iteration for the sequence $\{\mathbf{e}_{t}\}_{t}$. In the first case, the sample selected on the datasets $\mathcal{S}$ and $\mathcal{S}^{(i)}$ is identical, meaning that the data index $i_t\neq i$. Since the datasets containing $n$ samples have only one data that is different, the probability of this situation occurring is $1-1/n$. We directly calculate the gradients of the quadratic loss function, which evaluate models $\mathbf{w}$ and $\mathbf{w}'$ on the same data $\mathbf{z}_{i_t}$, and the following equation can be obtained.	
	\begin{equation}
		\nonumber
		\begin{split}
			&\mathbb{E}[\mathbf{e}_{t+1}]=\mathbb{E}[\mathbf{w}_{t+1}-\mathbf{w}_{t+1}']\\
			&=\mathbb{E}[\mathbf{w}_{t}-\mathbf{w}_{t}']-\eta\mathbb{E}\left[\nabla f(\mathbf{w}_{t-\tau}; \mathbf{z}_{i_t})-\nabla f(\mathbf{w}_{t-\tau}'; \mathbf{z}_{i_t})\right]\\
			&=\mathbb{E}[\mathbf{e}_{t}]-\eta\mathbb{E}\left[\mathbf{x}_{i_t}\mathbf{x}_{i_t}^{\top}(\mathbf{w}_{t-\tau}-\mathbf{w}_{t-\tau}')\right]\\
			&=\mathbb{E}[\mathbf{e}_{t}]-\eta \mathbb{E}[\mathbf{A} \mathbf{e}_{t-\tau}],
		\end{split}
	\end{equation}
	where the last equality is due to the fact that $\mathbf{z}_{i_t}$ is independent of $\mathbf{w}_{t-\tau}-\mathbf{w}_{t-\tau}'$ and the expectation on the training data set $\mathcal{S}$. The second case is that the algorithm exactly selects the $i$-th sample point which is different in the two datasets. Note that this happens only with probability $1/n$ in the random selection setup. In this case, we use the gradient containing noise to quantify the effect of different samples $\mathbf{z}_i$ and $\mathbf{z}_i'$ on the sequence.
	\begin{equation}
		\nonumber
		\begin{split}
			&\mathbb{E}[\mathbf{e}_{t+1}]=\mathbb{E}[\mathbf{w}_{t+1}-\mathbf{w}_{t+1}']\\
			&=\mathbb{E}[\mathbf{w}_{t}-\mathbf{w}_{t}']-\eta\mathbb{E}\left[\mathbf{A}\mathbf{w}_{t-\tau}+\mathbf{b}+\bm{\xi}_{t}-(\mathbf{A}\mathbf{w}_{t-\tau}'+\mathbf{b}+\bm{\xi}_{t}')\right]\\
			&=\mathbb{E}[\mathbf{e}_{t}]-\eta \mathbb{E}[\mathbf{A} \mathbf{e}_{t-\tau}]-\eta\mathbb{E}[\bm{\xi}_{t}-\bm{\xi}_{t}'].
		\end{split}
	\end{equation}
	Here, $\bm{\xi}_{t}'$ is the gradient noise with respect to $\nabla F_{\mathcal{S}}(\mathbf{w}_{t-\tau}')$ instead of $\nabla F_{\mathcal{S}^{(i)}}(\mathbf{w}_{t-\tau}')$. Since $\mathcal{S}$ and $\mathcal{S}^{(i)}$ differ by only one data point, the expected difference between $\nabla F_{\mathcal{S}}(\mathbf{w}_{t-\tau}')$ and $\nabla F_{\mathcal{S}^{(i)}}(\mathbf{w}_{t-\tau}')$ is of order $\mathcal{O}(1/n)$, hence $\bm{\xi}_{t}'$ also satisfies the bounded noise assumption. Combining the two cases and taking expectation for $\mathbf{e}_{t+1}$ with respect to the randomness of the algorithm, we derive the recurrence relation
	\begin{equation}
		\label{rec:e}
		\begin{split}
			\mathbb{E}[\mathbf{e}_{t+1}]&=\big(1-\frac{1}{n}\big)\mathbb{E}\big[\mathbf{e}_{t+1}|i_{t}\neq i, i_{t}\in[n]\big]+\frac{1}{n}\mathbb{E}[\mathbf{e}_{t+1}|i_{t}=i]\\
			&=\mathbb{E}[\mathbf{e}_{t}]-\eta \mathbb{E}[\mathbf{A} \mathbf{e}_{t-\tau}]-\frac{\eta}{n}\mathbb{E}[\bm{\xi}_{t}-\bm{\xi}_{t}'|i_{t}=i].
		\end{split}
	\end{equation}
	\begin{remark}
		\label{rmk:condition}
		For brevity, in all subsequent equational derivations, we abbreviate $\mathbb{E}[\bm{\xi}_{t}-\bm{\xi}_{t}'|i_{t}=i]$ as $\mathbb{E}[\bm{\xi}_{t}-\bm{\xi}_{t}']$. Furthermore, in the inequality derivations, we utilize Assumption \ref{amp:3} to bound $\mathbb{E}[\|\bm{\xi}_{t}-\bm{\xi}_{t}'\||i_{t}=i]$, i.e., $\mathbb{E}[\|\bm{\xi}_{t}-\bm{\xi}_{t}'\||i_{t}=i]\leq2\sigma$. This formulation does not affect our final results.
	\end{remark}
	For the sequence $\{\mathbf{s}_{t}\}_{t}$, we do not need to examine them separately. The iterative format \eqref{def:sgd} of delayed SGD produces the following equation.
	\begin{equation}
		\label{rec:r}
		\begin{split}
			&\mathbb{E}[\mathbf{s}_{t+1}]=\mathbb{E}[\mathbf{w}_{t+1}+\mathbf{w}_{t+1}']\\
			&=\mathbb{E}[\mathbf{w}_{t}+\mathbf{w}_{t}']-\eta\mathbb{E}\left[\mathbf{A}\mathbf{w}_{t-\tau}+\mathbf{b}+\bm{\xi}_{t}+(\mathbf{A}\mathbf{w}_{t-\tau}'+\mathbf{b}+\bm{\xi}_{t}')\right]\\
			&=\mathbb{E}[\mathbf{s}_{t}]-\eta \mathbb{E}[\mathbf{A} \mathbf{s}_{t-\tau}]-\eta\mathbb{E}[2\mathbf{b}+\bm{\xi}_{t}+\bm{\xi}_{t}'].
		\end{split}
	\end{equation}
	
	The well-established equational recurrence formulas for sequences $\{\mathbf{e}_{t}\}_{t}$ and $\{\mathbf{s}_{t}\}_{t}$ motivate us to study them using the \textit{generating function} tool. Generally speaking, the generating function of a given sequence $\{\bm{\alpha}_t\}_t$ is a formal power series $\bm{\varphi}(x)=\sum_{t=0}^\infty \bm{\alpha}_t x^t$. Note that the variable $x$ in generating functions serves as a placeholder to track coefficients without standing for any particular value. We define $[x^{t}]\bm{\varphi}(x)$ as the operation that extracts the coefficient of $x^{t}$ in the formal power series $\bm{\varphi}(x)$, i.e., $\bm{\alpha}_t=[x^{t}]\bm{\varphi}(x)$. Generating functions are frequently the most efficient and concise way to present information about their coefficients, and have been used to study optimization errors of delayed gradient methods \cite{arjevani2020tight}. Details on generating functions can be found in \cite{WILF199027, stanley_fomin_1999, flajolet_sedgewick_2009}. 
	\begin{remark}
		\label{rmk:01}
		Arjevani et al. \cite{arjevani2020tight} investigated the optimization error, i.e., $F_{\mathcal{S}}(\mathbf{w}_T)-F_{\mathcal{S}}(\mathbf{w}^*)$, of the delayed SGD algorithm, where $\mathbf{w}^*$ is the minimizer of $F_{\mathcal{S}}$. They used the crucial property $\nabla F_{\mathcal{S}}(\mathbf{w}^*)=\bm{0}$ to construct the equational recursive sequence so that it can be analyzed with the generating functions. This paper, however, studies the generalization error, i.e., $F(\mathbf{w}_T)-F_{\mathcal{S}}(\mathbf{w}_T)$, which is not relevant to $\mathbf{w}^*$, and thus is quite different from \cite{arjevani2020tight} in constructing the equational recursive relations. We ingeniously formulate different formats of the delayed stochastic gradient, denoted as \eqref{grad_format}, and employ them flexibly to derive the average stability, yielding equational recurrence formulas \eqref{rec:e} and \eqref{rec:r} for sequences $\{\mathbf{e}_{t}\}_{t}$ and $\{\mathbf{s}_{t}\}_{t}$, respectively.
	\end{remark}
	Let $\bm{\phi}(x)$ and $\bm{\psi}(x)$ be the generating functions of sequences $\{\mathbf{e}_{t}\}_{t}$ and $\{\mathbf{s}_{t}\}_{t}$, respectively, defined as
	\begin{equation}
		\nonumber
		\bm{\phi}(x)=\sum_{t=0}^\infty \mathbf{e}_t x^t ~~ \text{and} ~~ \bm{\psi}(x)=\sum_{t=0}^\infty \mathbf{s}_t x^t.
	\end{equation}
	The expectation notation is omitted in the derivation of the generating function for the sake of brevity. According to recurrence formula \eqref{rec:e}, the generating function $\bm{\phi}(x)$ proceeds as
	\begin{equation}
		\label{rec:phi}
		\begin{split}
			&\bm{\phi}(x) =\sum_{t=0}^{\tau}\mathbf{e}_{t} x^t+\sum_{t=\tau+1}^{\infty}\mathbf{e}_{t} x^t\\
			&\overset{(\star)}{=}\sum_{t=\tau+1}^{\infty}\left[\mathbf{e}_{t-1}-\eta \mathbf{A} \mathbf{e}_{t-\tau-1}-\frac{\eta}{n}(\bm{\xi}_{t-1}-\bm{\xi}_{t-1}')\right] x^t\\
			&\overset{(\star)}{=}x\bm{\phi}(x)-\eta \mathbf{A}x^{\tau+1} \bm{\phi}(x)-\frac{\eta}{n}\sum_{t=\tau}^{\infty}(\bm{\xi}_{t}-\bm{\xi}_{t}')x^{t+1},
		\end{split}
	\end{equation}
	where $(\star)$ uses the initialization \eqref{init}. Similarly, the generating function $\bm{\psi}(x)$ can deduce the following property based on equation \eqref{rec:r}.
	\begin{equation}
		\label{rec:psi}
		\begin{split}
			&\bm{\psi}(x)=\sum_{t=0}^\tau \mathbf{s}_t x^t\\
			&\qquad+\sum_{t=\tau+1}^{\infty}\Big[\mathbf{s}_{t-1}-\eta \mathbf{A} \mathbf{s}_{t-\tau-1}-\eta(2\mathbf{b}+\bm{\xi}_{t-1}+\bm{\xi}_{t-1}')\Big] x^t\\
			&\overset{(\star)}{=}\mathbf{s}_0+x\bm{\psi}(x)-\eta \mathbf{A} x^{\tau+1}\bm{\psi}(x)-\eta \sum_{t=\tau}^{\infty}(2\mathbf{b}+\bm{\xi}_{t}+\bm{\xi}_{t}')x^{t+1}.
		\end{split}
	\end{equation}
	Abbreviating the last terms in equations \eqref{rec:phi} and \eqref{rec:psi} as the power series $\mathbf{u}(x)$ and $\mathbf{v}(x)$, i.e.,
	\begin{equation}
		\nonumber
		\mathbf{u}(x)=\sum_{t=\tau}^{\infty}(\bm{\xi}_{t}'-\bm{\xi}_{t})x^{t+1} ~~ \text{and} ~~ \mathbf{v}(x)=\sum_{t=\tau}^{\infty}(2\mathbf{b}+\bm{\xi}_{t}+\bm{\xi}_{t}')x^{t+1}.
	\end{equation}
	Rearranging the terms in equations \eqref{rec:phi} and \eqref{rec:psi}, we arrive at
	\begin{equation}
		\nonumber
		\begin{split}
			(\mathbf{I}-\mathbf{I}x+\eta \mathbf{A} x^{\tau+1})\bm{\phi}(x)&=\frac{\eta}{n}\mathbf{u}(x) \\
			(\mathbf{I}-\mathbf{I}x+\eta \mathbf{A} x^{\tau+1})\bm{\psi}(x)&=\mathbf{s}_0-\eta \mathbf{v}(x),
		\end{split}
	\end{equation}
	where $\mathbf{I}\in\mathbb{R}^{d\times d}$ is the identity matrix. Below we introduce some algebraic notations for studying generating functions. $\mathbb{R}[[x]]$ represents the set of all formal power series in the indeterminate $x$ with coefficients belonging to the reals $\mathbb{R}$. It is equipped with standard addition and multiplication (Cauchy product) operations, forming a commutative ring. The set of all matrices in $\mathbb{R}^{d\times d}$ is denoted as $\mathcal{M}_{d}(\mathbb{R})$, and $\mathcal{M}_{d}(\mathbb{R})[[x]]$ is the set of matrices with elements in $\mathbb{R}[[x]]$, which also forms a ring of formal power series with real matrix coefficients. According to \cite[Proposition 2.1]{WILF199027}, $\mathbf{I}-\mathbf{I}x+\eta \mathbf{A} x^{\tau+1}$ is invertible in $\mathcal{M}_{d}(\mathbb{R})[[x]]$, as the constant term $\mathbf{I}$ is invertible in $\mathcal{M}_{d}(\mathbb{R})$. Denote the inverse of $\mathbf{I}-\mathbf{I}x+\eta \mathbf{A} x^{\tau+1}$ as the power series $\bm{\pi}(x)$, namely
	\begin{equation}
		\label{def:pi}
		\bm{\pi}(x)=(\mathbf{I}-\mathbf{I}x+\eta \mathbf{A} x^{\tau+1})^{-1},
	\end{equation}
	then the following concise properties are observed for the generating functions $\bm{\phi}(x)$ and $\bm{\psi}(x)$.
	\begin{equation}
		\nonumber
		\bm{\phi}(x)=\frac{\eta}{n}\bm{\pi}(x)\cdot\mathbf{u}(x) ~~ \text{and} ~~ \bm{\psi}(x)=\bm{\pi}(x)\cdot\left[\mathbf{s}_0-\eta \mathbf{v}(x)\right].
	\end{equation}	
	
	Returning to the average stability \eqref{avg_stab} of the delayed SGD algorithm, we can extract the corresponding coefficients in generating functions, i.e., $\mathbf{e}_{t}=[x^{t}]\bm{\phi}(x), \mathbf{s}_{t}=[x^{t}]\bm{\psi}(x)$, to analyze the terms in \eqref{avg_stab}. According to Lemma \ref{lem:pi}, we can obtain the following properties without considering expectations.
	\begin{equation}
		\label{pq}
		\|[x^{i}]\bm{\pi}(x)\|\leq p_{i}, ~ \big\|\sqrt{\mathbf{A}}[x^{i}]\bm{\pi}(x)\big\| \leq q_{i}, ~~ i\in\{0, 1, \ldots, t\},
	\end{equation}
	where $p_{i}, q_{i}\geq0$ are constants. We can then deduce that
	\begin{equation}
		\nonumber
		\begin{split}
			&\mathbb{E}\|\mathbf{b}\|\|\mathbf{e}_{t}\|=\frac{\eta}{n}\mathbb{E}\|\mathbf{b}\|\|[x^{t}]\big(\bm{\pi}(x)\cdot\mathbf{u}(x)\big)\|\\
			&\overset{(i)}{=}\frac{\eta}{n}\mathbb{E}\big\|\mathbf{b}\big\|\Big\|\sum_{i=0}^{t-\tau-1}\left([x^{i}]\bm{\pi}(x)\right) (\bm{\xi}_{t-i-1}'-\bm{\xi}_{t-i-1})\Big\|\\
			&\overset{(ii)}{\leq}\frac{\eta}{n}\sum_{i=0}^{t-\tau-1}p_{i}\mathbb{E}\big\|\mathbf{b}\big\|\big\|\bm{\xi}_{t-i-1}'-\bm{\xi}_{t-i-1}\big\|\overset{(iii)}{\leq}\frac{2\eta r\sigma}{n}\sum_{i=0}^{t-\tau-1}p_{i},
		\end{split}
	\end{equation}
	where $(i)$ is based on the Cauchy product for formal power series and the formulation of $\mathbf{u}(x)$, $(ii)$ utilizes the triangle, Cauchy-Schwartz inequalities along with \eqref{pq}, and $(iii)$ is followed by Assumptions \ref{amp:2}, \ref{amp:3}. Analogously, we have
	\begin{equation}
		\nonumber
		\begin{split}
			&\mathbb{E}\big\|\sqrt{\mathbf{A}}\mathbf{e}_{t}\big\|\big\|\sqrt{\mathbf{A}}\mathbf{s}_{t}\big\|=\mathbb{E}\big\|\sqrt{\mathbf{A}}[x^{t}]\big(\bm{\phi}(x)\big)\big\|\big\|\sqrt{\mathbf{A}}[x^{t}]\big(\bm{\psi}(x)\big)\big\|\\
			&\overset{(i)}{\leq}\frac{\eta}{n}\mathbb{E}\big\|\sqrt{\mathbf{A}}[x^{t}]\big(\bm{\pi}(x)\cdot\mathbf{u}(x)\big)\big\|\cdot\Big(\big\|\sqrt{\mathbf{A}}[x^{t}]\bm{\pi}(x)\mathbf{s}_0\big\|\\
			&\quad+\eta\big\|\sqrt{\mathbf{A}}[x^{t}]\big(\bm{\pi}(x)\cdot\mathbf{v}(x)\big)\big\|\Big)\\
			&\overset{(ii)}{\leq}\frac{\eta}{n}\mathbb{E}\Big\|\sum_{i=0}^{t-\tau-1}\big(\sqrt{\mathbf{A}}[x^{i}]\bm{\pi}(x)\big)(\bm{\xi}_{t-i-1}'-\bm{\xi}_{t-i-1})\Big\|\\
			&\quad\cdot\Big(\eta\Big\|\sum_{i=0}^{t-\tau-1}\big(\sqrt{\mathbf{A}}[x^{i}]\bm{\pi}(x)\big)(2\mathbf{b}+\bm{\xi}_{t-i-1}'+\bm{\xi}_{t-i-1})\Big\|\\
			&\qquad+\big\|\sqrt{\mathbf{A}}[x^{t}]\bm{\pi}(x)\mathbf{s}_0\big\|\Big)\\
			&\overset{(iii)}{\leq}\frac{4\eta\sigma\|\mathbf{w}_0\|}{n}q_{t}\sum_{i=0}^{t-\tau-1}q_{i}+\frac{4\eta^{2}\sigma(r+\sigma)}{n} \Big(\sum_{i=0}^{t-\tau-1}q_{i}\Big)^{2},
		\end{split}
	\end{equation}
	where $(i)$ uses the linearity of the extraction operation $[x^{t}]$, and $(ii)$ follows by the Cauchy product for formal power series. $(iii)$ utilizes inequality \eqref{pq}, initialization $\mathbf{s}_0=2\mathbf{w}_0$, and Assumptions \ref{amp:2}, \ref{amp:3}. Finally, we derive the following average stability bound of delayed SGD.
	\begin{proposition}
		\label{thm:stab}
		Let Assumptions \ref{amp:2}, \ref{amp:3} hold and the loss function $F_{\mathcal{S}}$ be quadratic, i.e., $F_{\mathcal{S}}(\mathbf{w})=\frac{1}{2}\mathbf{w}^{\top}\mathbf{A} \mathbf{w}+\mathbf{b}^{\top}\mathbf{w}+c$, then the average stability $\epsilon_{\text{stab}}$ of delayed SGD after $t$ iterations ($t>\tau$) satisfies
		\begin{equation}
			\label{avg_stab_sgd}
			\begin{split}
				\epsilon_{\text{stab}}\leq&\frac{2\eta\sigma\|\mathbf{w}_0\|}{n}q_{t}\sum_{i=0}^{t-\tau-1}q_{i}+\frac{2\eta^{2}\sigma(r+\sigma)}{n} \Big(\sum_{i=0}^{t-\tau-1}q_{i}\Big)^{2}\\
				&+\frac{2\eta r\sigma}{n}\sum_{i=0}^{t-\tau-1}p_{i},
			\end{split}
		\end{equation}
		where $p_{i}, q_{i}\geq0$ are constants defined in \eqref{pq}.
	\end{proposition}

	\section{Generalization error of delayed stochastic gradient descent}
	\label{sec:gen}
	According to Lemma \ref{lem:1} and the average stability \eqref{avg_stab_sgd}, we need to estimate the coefficients of the power series $\bm{\pi}(x)$ to determine the constants $p_{i}, q_{i}$ in inequality \eqref{pq}. Inspired by \cite[Lemma 1]{arjevani2020tight}, we establish the subsequent lemma to bound $\|[x^{t}]\bm{\pi}(x)\|$ and $\|\sqrt{\mathbf{A}}[x^{t}]\bm{\pi}(x)\|$. Then, we are ready to present the generalization error of delayed SGD, including the convex and strongly convex quadratic objectives. All the proof details are included in the Supplementary materials.
	\begin{lemma}
		\label{lem:pi}
		Let Assumption \ref{amp:1} holds and the learning rate $\eta \in (0, 1/\mu\tau]$. Then the $t$-th coefficient of the power series $\bm{\pi}(x)$ satisfies $\|[x^{t}]\bm{\pi}(x)\|\leq 1$ for any $t\geq0$. Furthermore, if $\eta \in (0, 1/20\mu(\tau+1)]$, we have
		\begin{equation}
			\nonumber
			\left\{
			\begin{array}{ll}\vspace{1.1ex}
				\left\|[x^{t}]\bm{\pi}(x)\right\|  \leq 1   & ~~ 0 \leq t \leq t_0-1, \\ 
				\left\|[x^{t}]\bm{\pi}(x)\right\| \leq 3\max_{j\in[d]}(1-\eta a_{j})^{t+1} & ~~ t \geq t_0.
			\end{array}\right.
		\end{equation}
		\begin{equation}
			\nonumber
			\left\{
			\begin{array}{ll}\vspace{1.1ex}
				\big\|\sqrt{\mathbf{A}}[x^{t}]\bm{\pi}(x)\big\| \leq \max_{j\in[d]}\sqrt{a_{j}}    \qquad\qquad ~~ 0 \leq t \leq t_0-1, \\ 
				\big\|\sqrt{\mathbf{A}}[x^{t}]\bm{\pi}(x)\big\| \leq 3\max_{j\in[d]}\sqrt{a_{j}} (1-\eta a_{j})^{t+1} \qquad t \geq t_0.
			\end{array}\right.
		\end{equation}
		where $t_0=(\tau+1)\ln(2(\tau+1))$ and $a_{j}$ is the $j$-th eigenvalue of the positive semi-definite matrix $\mathbf{A}$.
	\end{lemma}
	\begin{theorem}
		\label{thm:1}
		Let Assumptions \ref{amp:1}-\ref{amp:3} hold, and the learning rate $\eta \in (0, 1/20\mu(\tau+1)]$, then the generalization error (or average stability) of delayed SGD after $T$ iterations satisfies
		\begin{equation}
			\nonumber
			\begin{split}
				&\mathbb{E}_{\mathcal{S}, \mathcal{A}}[F(\mathbf{w}_{T})-F_{\mathcal{S}}(\mathbf{w}_{T})]\\
				&\leq\frac{\sigma(r+\sigma)}{n\mu}\left[\sqrt{T-\tau}+12\mu\left\|\mathbf{w}_0\right\|+\ln^{2}(\tau+1)\right]\\
				&\quad+\frac{\sigma(r+\sigma)}{n\mu\tau}(T-\tau)\\
				&\leq\widetilde{\mathcal{O}}\left(\frac{T-\tau}{n\tau}\right).
			\end{split}
		\end{equation}
	\end{theorem}
	For the quadratic problem, Theorem \ref{thm:1} indicates that the algorithmic stability of delayed SGD is positively correlated with the delay $\tau$ provided the learning rate satisfies $\eta\leq 1/20\mu(\tau+1)$, meaning that asynchronous delay makes SGD more stable and hence reduces the generalization error. It follows that asynchronous delay enables the model to perform more consistently across the training and test datasets. The next theorem demonstrates that the generalization error is independent of the iteration number $T$ if the quadratic function is further strongly convex.
	\begin{theorem}
		\label{thm:2}
		Suppose the quadratic function is $\lambda$-strongly convex, and Assumptions \ref{amp:1}-\ref{amp:3} hold. Let the learning rate $\eta \in (0, 1/20\mu(\tau+1)]$, then the generalization error of delayed SGD is bounded by
		\begin{equation}
			\nonumber
			\begin{split}
				&\mathbb{E}_{\mathcal{S}, \mathcal{A}}[F(\mathbf{w}_{T})-F_{\mathcal{S}}(\mathbf{w}_{T})]\\
				&\leq\frac{2\sigma(r+\sigma)\eta t_{0}}{n}\left[1+3\mu\left\|\mathbf{w}_0\right\|+\mu\eta t_0+12\sqrt{\frac{\mu}{e\lambda}}\right]\\
				&\quad+\frac{42\sigma(r+\sigma)}{n\lambda}+\frac{36\sigma\left\|\mathbf{w}_0\right\|}{n}\sqrt{\frac{\mu}{e\lambda}}\\
				&\leq\widetilde{\mathcal{O}}\left(\frac{1}{n}\right).
			\end{split}
		\end{equation}
	\end{theorem}
	Additionally, both Theorems \ref{thm:1} and \ref{thm:2} suggest that increasing the training samples amount $n$ can stabilize the algorithm and thus decrease the generalization error.
	\begin{remark}
		Under the bounded gradient assumption, \cite{hardt2016train} provides upper bounds on the generalization error of plain SGD for both general convex and strongly convex cases as $\mathcal{O}(\frac{T}{n})$ and $\mathcal{O}(\frac{1}{n})$, respectively, and these upper bounds are subsequently shown to be tight \cite{zhang2022stability}. This is consistent with our results in the synchronous case (i.e., $\tau=1$).
	\end{remark}
	\begin{remark}
		It is worth mentioning that the proven results also benefit the understanding of delayed or synchronous algorithms for deep neural network (DNN) training though they are established for quadratic functions. The reason lies in the neural tangent kernel (NTK) approach \cite{jacot2018neural, NEURIPS2019_0d1a9651}, which promises that the dynamics of gradient descent on DNNs are close to those on quadratic optimization under sufficient overparameterization and random initialization.		
	\end{remark}
	\begin{remark}
		Denote $\mathbf{w}_*$ as the minimizer of population risk $F$, and $F(\mathbf{w}_{T})-F(\mathbf{w}_{*})$ is known as the \textit{excess generalization error}, which is bounded by the sum of the generalization error and the optimization error \cite{hardt2016train}. Together with the optimization results of \cite{arjevani2020tight}, we can derive upper bounds for the excess generalization error of delayed SGD, namely, $\widetilde{\mathcal{O}}(\frac{1}{\sqrt{T}}+\frac{T-\tau}{n\tau})$ for convex and $\widetilde{\mathcal{O}}(\frac{1}{T}+\frac{1}{n})$ for strongly convex quadratic loss function. The results suggest that when solving convex quadratic problems, training should be stopped at the proper time, i.e., choosing an appropriate $T$ to balance the optimization and generalization errors so that the excess generalization error is minimized. While in the strongly convex case, sufficient training can be performed to reduce optimization errors and improve the generalization performance.
	\end{remark}
	
	\begin{figure*}[!t]
		\centering
		\subfigure[Gen. error of \texttt{rcv1}]{
			\label{fig:rcv1_fixed}
			\includegraphics[width=0.24\linewidth]{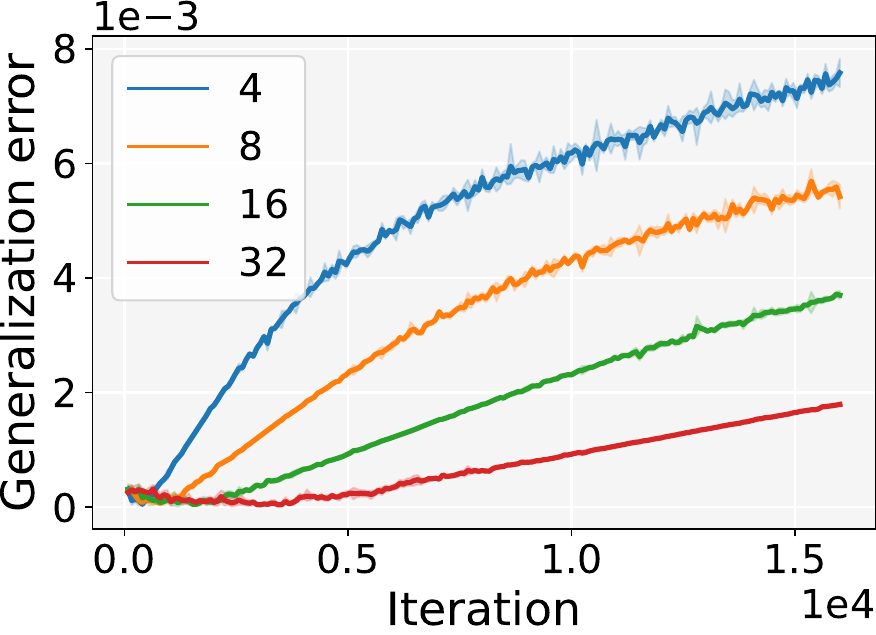}}
		\subfigure[Gen. error of \texttt{gisette}]{
			\label{fig:gisette_fixed}
			\includegraphics[width=0.24\linewidth]{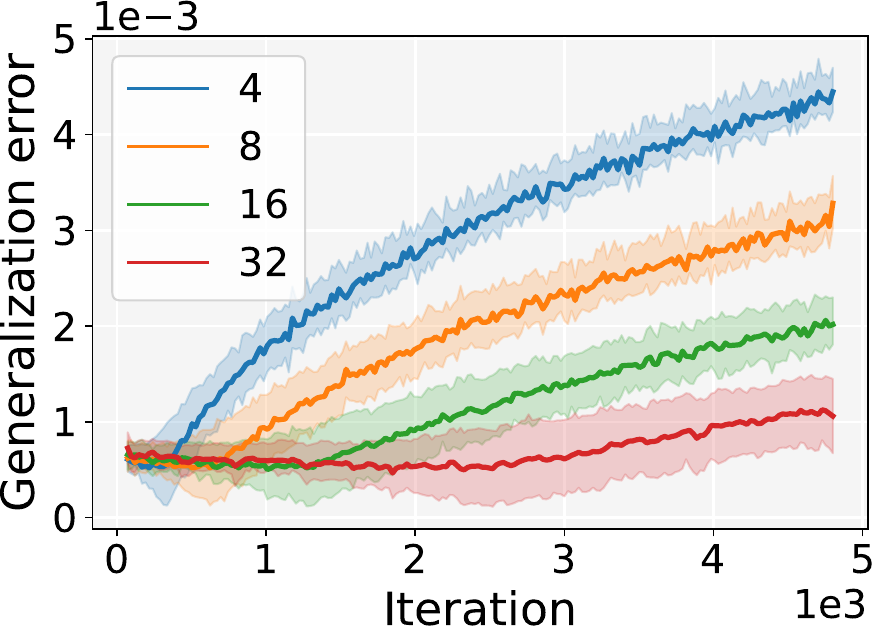}}
		\subfigure[Gen. error of \texttt{covtype}]{
			\label{fig:covtype_fixed}
			\includegraphics[width=0.24\linewidth]{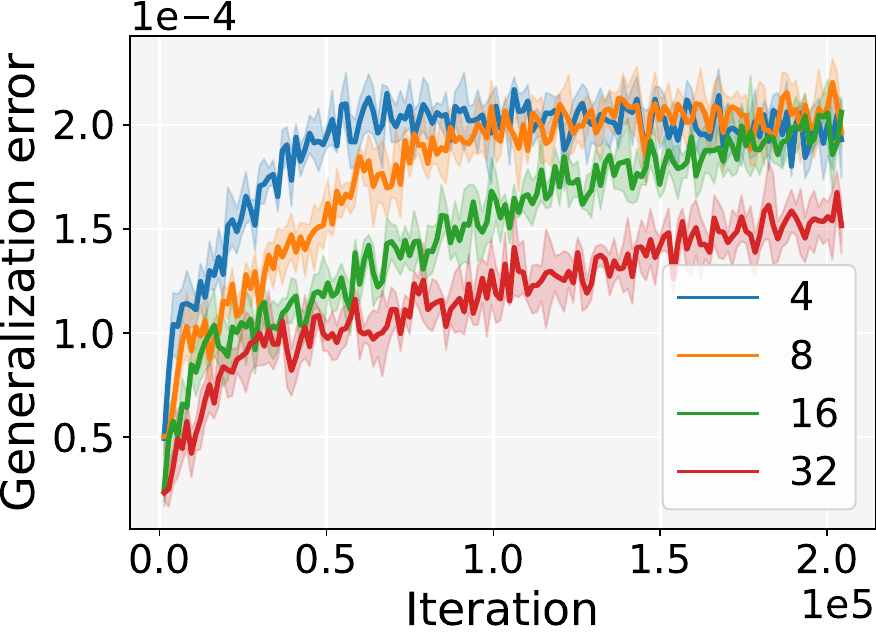}}
		\subfigure[Gen. error of \texttt{ijcnn1}]{
			\label{fig:ijcnn1_fixed}
			\includegraphics[width=0.24\linewidth]{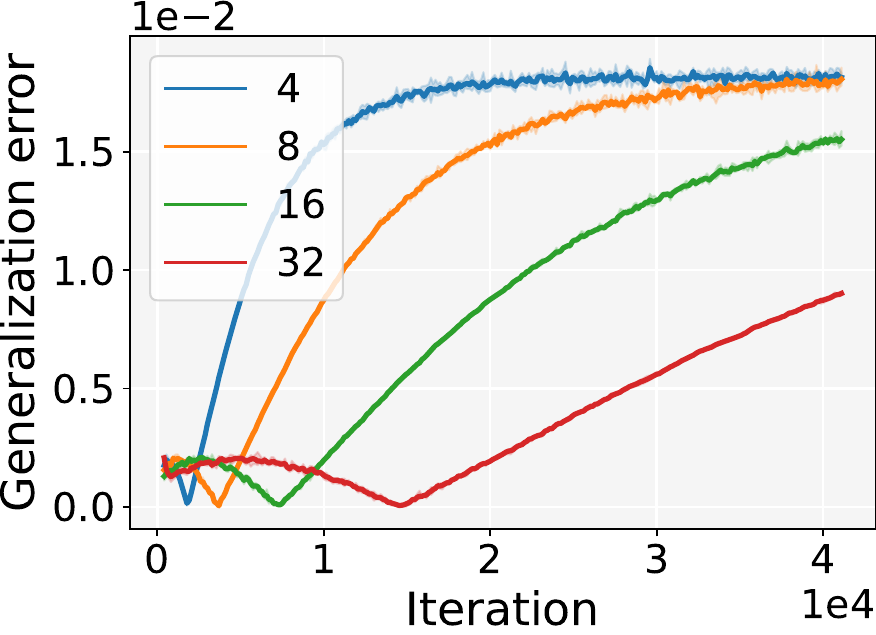}}
		\caption{Experimental results for solving quadratic convex problems with delayed SGD. We test the generalization error of different \textit{fixed} delays on four LIBSVM datasets \texttt{rcv1}, \texttt{gisette}, \texttt{covtype} and \texttt{ijcnn1}. The variance in the plots is due to selecting different random seeds in multiple trials.}
		\label{fig:gen_fixed}
	\end{figure*}
	
	\begin{figure*}[!t]
		\centering
		\subfigure[Gen. error of \texttt{rcv1}]{
			\label{fig:rcv1_random}
			\includegraphics[width=0.24\linewidth]{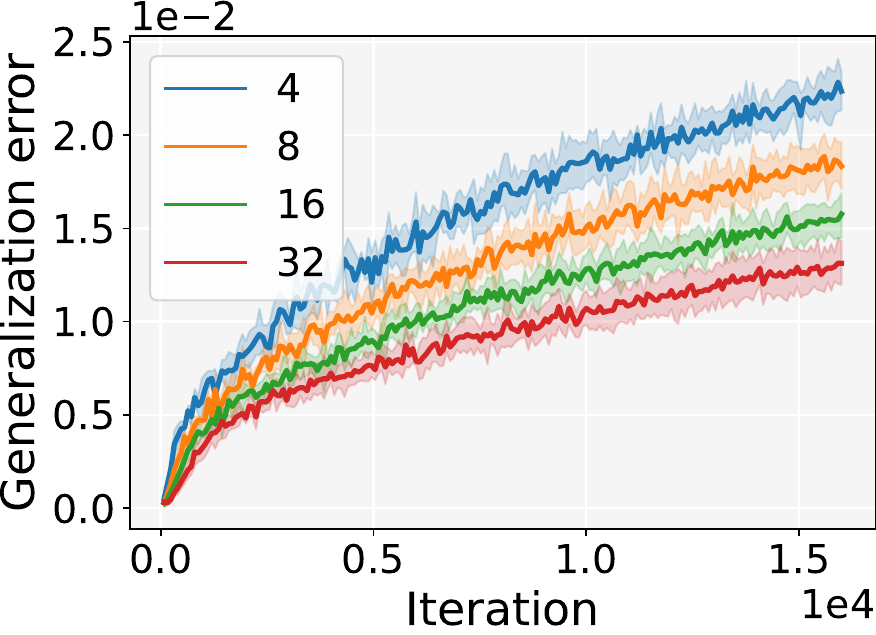}}
		\subfigure[Gen. error of \texttt{gisette}]{
			\label{fig:gisette_random}
			\includegraphics[width=0.24\linewidth]{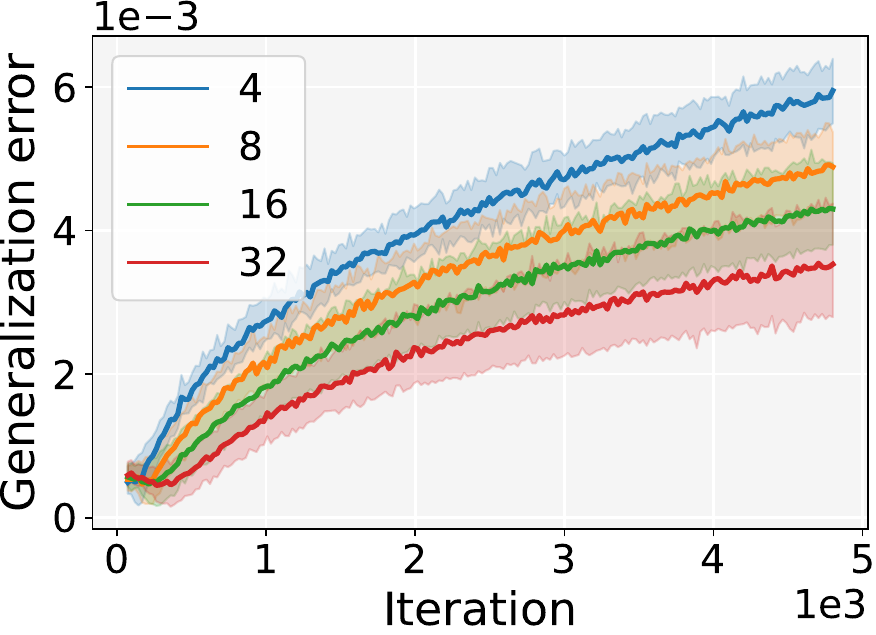}}
		\subfigure[Gen. error of \texttt{covtype}]{
			\label{fig:covtype_random}
			\includegraphics[width=0.24\linewidth]{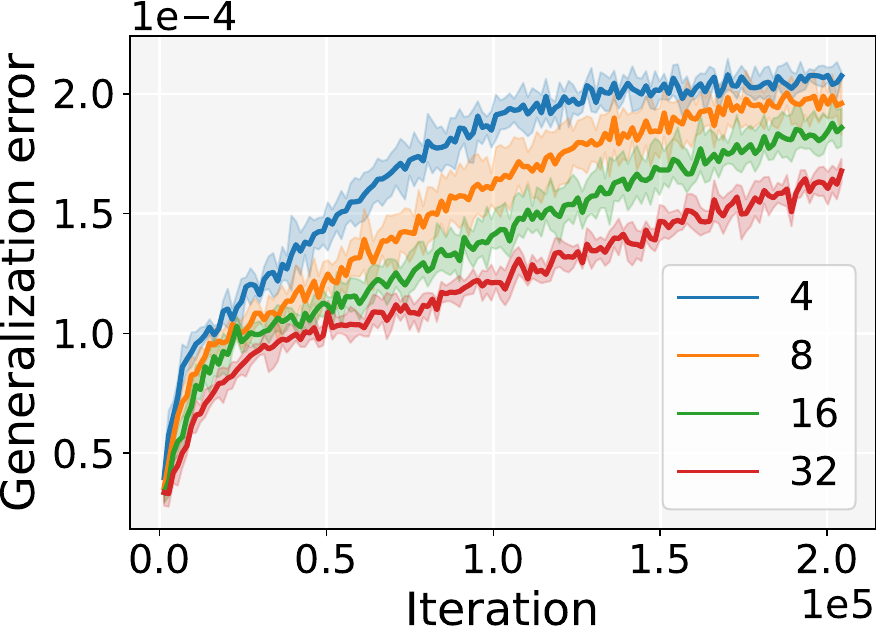}}
		\subfigure[Gen. error of \texttt{ijcnn1}]{
			\label{fig:ijcnn1_random}
			\includegraphics[width=0.24\linewidth]{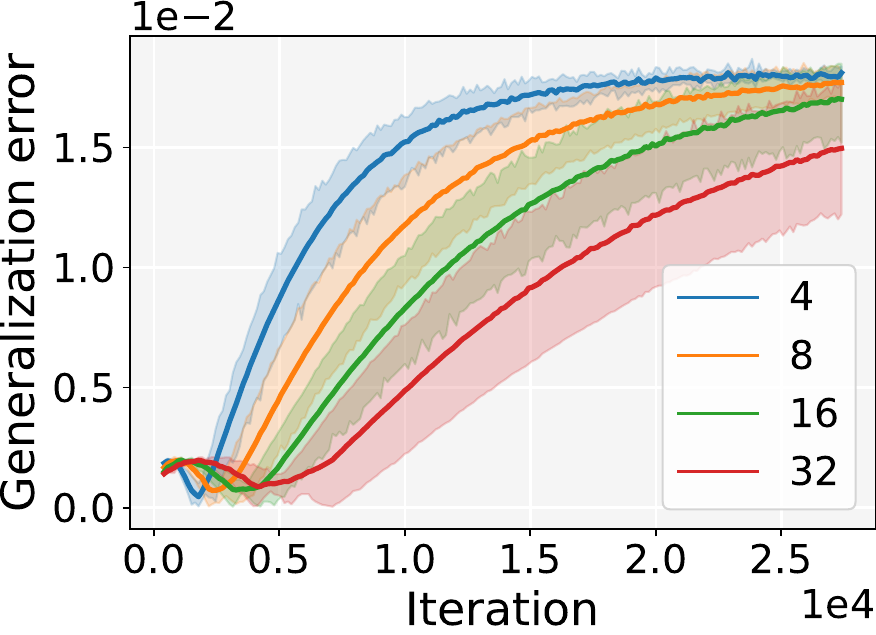}}
		\caption{Generalization error of solving quadratic convex problems by the delayed SGD algorithm with \textit{random} delays.}
		\label{fig:gen_random}
	\end{figure*}

	\section{Extension to random delays}
	\label{sec:random_delay}
	This section shows how to extend the fixed delay $\tau$ to a dynamic scenario with the random delay $\tau_t$ for stochastic gradient descent. Specifically, the algorithm with random delay is formulated as 
	\begin{equation}
		\nonumber
		\begin{split}
			\mathbf{w}_{t+1}&=\mathbf{w}_{t}-\eta \nabla f(\mathbf{w}_{t-\tau_t}; \mathbf{z}_{i_t})\\
			&=\mathbf{w}_{t}-\eta\left(\nabla F_{\mathcal{S}}(\mathbf{w}_{t-\tau_t})+\bm{\xi}_{t}\right),
		\end{split}
	\end{equation}
	where $\tau_{t}$ is the random delay at $t$-th iteration, and the standard bounded delay assumption is required.
	\begin{assumption}[bounded delay]
		\label{amp:4}
		The random delay is bounded, i.e., there exists a constant $\overline{\tau}\in\mathbb{R}$ such that $\tau_{t}\leq\overline{\tau}$ for all iteration $t$.
	\end{assumption}
	To start the training process, we initialize the delayed algorithm as $\mathbf{w}_{0}=\cdots=\mathbf{w}_{\overline{\tau}}$. Then we follow the same notations $\mathbf{e}_{t}=\mathbf{w}_{t}-\mathbf{w}_{t}'$ and $\mathbf{s}_{t}=\mathbf{w}_{t}+\mathbf{w}_{t}'$ in Section \ref{sec:stab}. The equational recurrence formulas on sequences $\{\mathbf{e}_{t}\}_{t}$ and $\{\mathbf{s}_{t}\}_{t}$ are identical to \eqref{rec:e} and \eqref{rec:r}, except that the fixed delay $\tau$ is replaced by the random delay $\tau_{t}$. The related generating functions, $\bm{\phi}(x)$ and $\bm{\psi}(x)$, are determined as
	\begin{equation}
		\nonumber
		\begin{split}
			\bm{\phi}(x)&=x\bm{\phi}(x)-\eta \mathbf{A} x^{\overline{\tau}+1} \bm{\phi}(x)\\
			&\quad-\eta\sum_{t=\overline{\tau}}^{\infty}\Big[\frac{1}{n}(\bm{\xi}_{t}-\bm{\xi}_{t}')+\bm{\zeta}_{t}-\bm{\zeta}_{t}'\Big]x^{t+1},\\
			\bm{\psi}(x)&=\mathbf{s}_0+x\bm{\psi}(x)-\eta \mathbf{A} x^{\overline{\tau}+1} \bm{\psi}(x)\\
			&\quad-\eta \sum_{t=\overline{\tau}}^{\infty}(2\mathbf{b}+\bm{\xi}_{t}+\bm{\xi}_{t}'+\bm{\zeta}_{t}+\bm{\zeta}_{t}')x^{t+1},
		\end{split}
	\end{equation}
	where $\bm{\zeta}_{t}$ and $\bm{\zeta}_{t}'$ measure the magnitude of model differences $\mathbf{w}_{t-\tau_t}-\mathbf{w}_{t-\overline{\tau}}$ and $\mathbf{w}_{t-\tau_t}'-\mathbf{w}_{t-\overline{\tau}}'$, respectively. Due to the convergence property of delayed SGD \cite{mishchenko2022asynchronous} and the bounded delay Assumption \ref{amp:4}, it is reasonable to conclude that $\bm{\zeta}_{t}$ and $\bm{\zeta}_{t}'$ are bounded. By extracting the corresponding coefficients of the generating functions to analyze average stability \eqref{avg_stab}, we can establish the generalization error bounds of the delayed SGD algorithm with random delay.
	\begin{corollary}
		Let Assumptions \ref{amp:1}-\ref{amp:4} hold, and the learning rate $\eta \in (0, 1/20\mu(\overline{\tau}+1)]$, then we can derive the generalization error bounds of delayed SGD with random delays for the convex case $\widetilde{\mathcal{O}}(\frac{T-\overline{\tau}}{n\overline{\tau}})$ and the strongly convex case $\widetilde{\mathcal{O}}(\frac{1}{n})$, respectively.
	\end{corollary}
	
	\begin{figure*}[!t]
		\centering
		\subfigure[Train loss of fixed delays]{
			\label{fig:loss1_fixd}
			\includegraphics[width=0.24\linewidth]{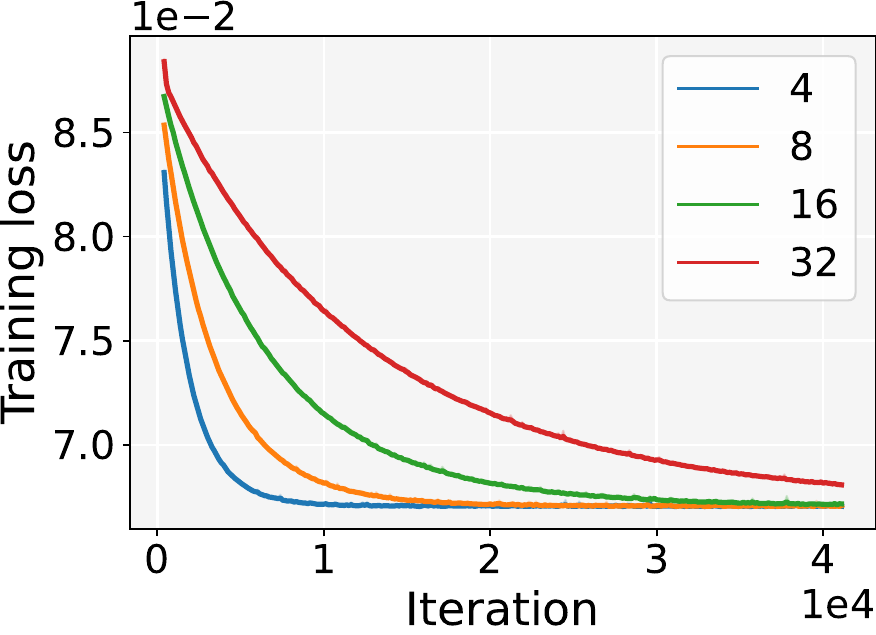}}
		\subfigure[Test loss of fixed delays]{
			\label{fig:loss2_fixd}
			\includegraphics[width=0.24\linewidth]{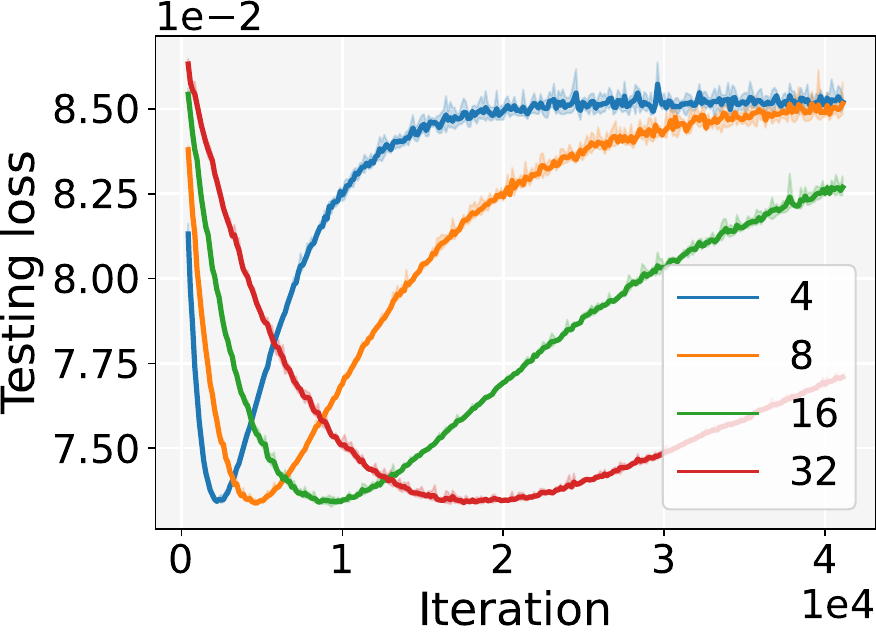}}
		\subfigure[Train loss of random delays]{
			\label{fig:loss1_random}
			\includegraphics[width=0.24\linewidth]{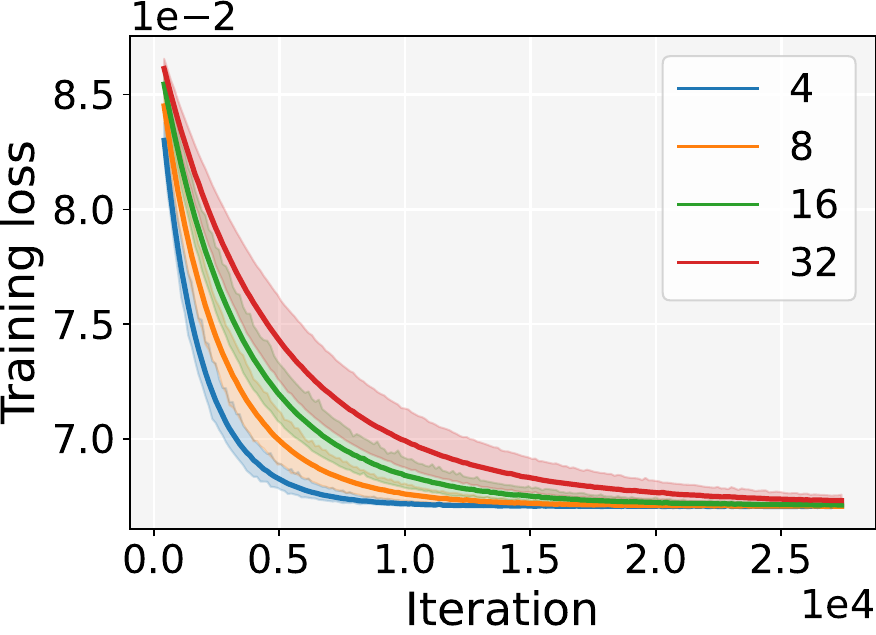}}
		\subfigure[Test loss of random delays]{
			\label{fig:loss2_random}
			\includegraphics[width=0.24\linewidth]{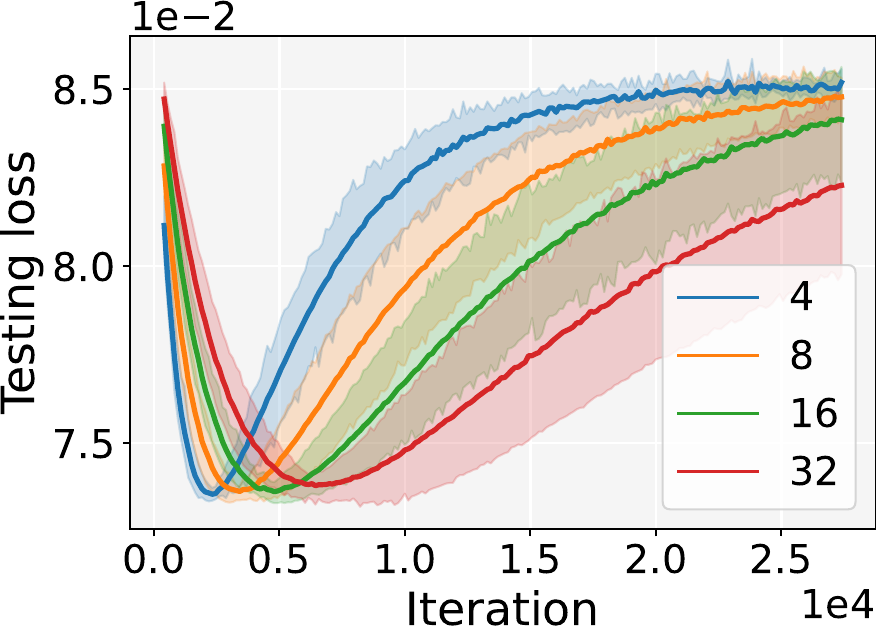}}
		\caption{Training and testing loss of solving quadratic convex problems by the delayed SGD algorithm with fixed and random delays.}
		\label{fig:loss}
	\end{figure*}
	
	\begin{figure*}[htbp]
		\centering
		\subfigure[Gen. error of MNIST]{
			\label{fig:app-mnist}
			\includegraphics[width=0.235\linewidth]{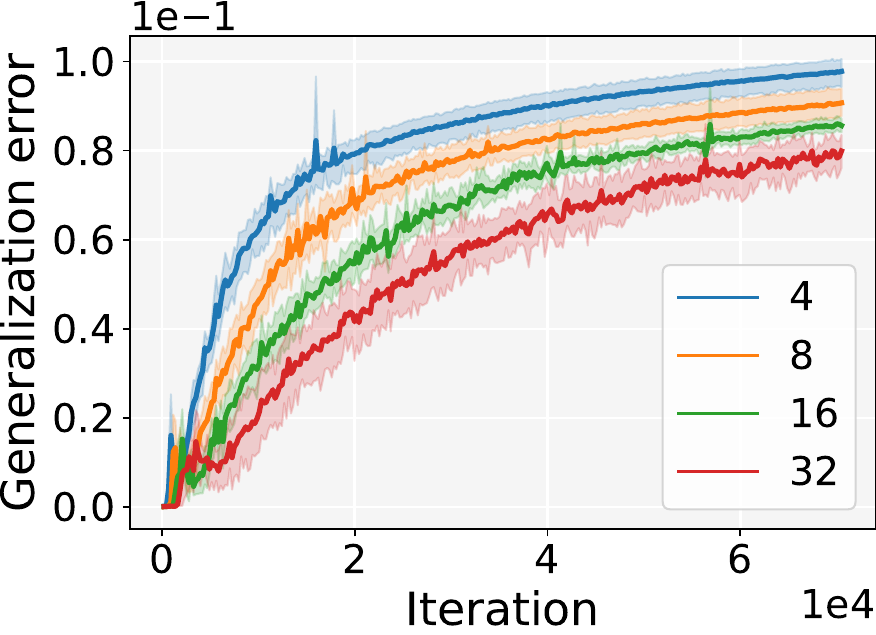}}
		\subfigure[Gen. error of CIFAR-10]{
			\label{fig:app-cifar10}
			\includegraphics[width=0.235\linewidth]{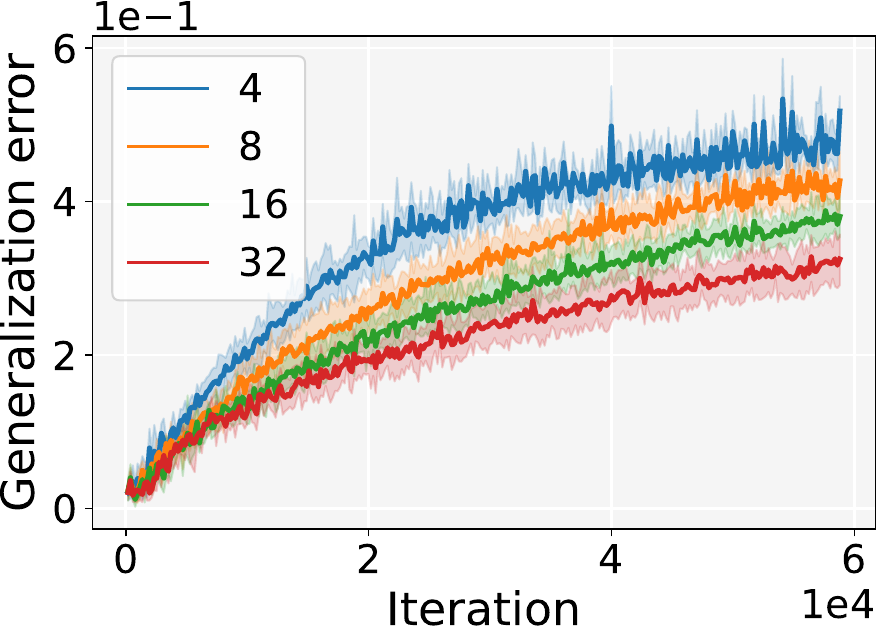}}
		\subfigure[Training loss of CIFAR-100]{
			\label{fig:app-cifar100-train}
			\includegraphics[width=0.235\linewidth]{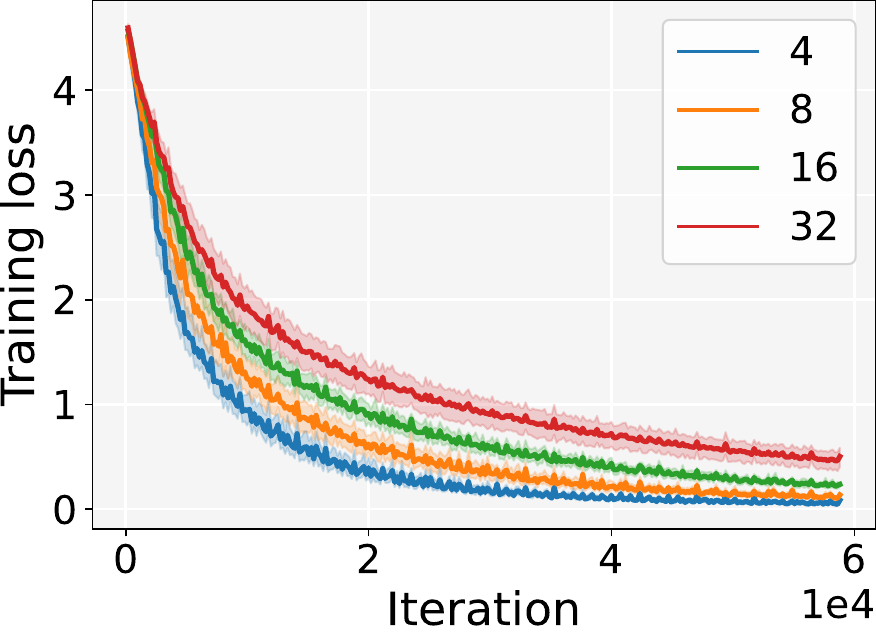}}
		\subfigure[Testing loss of CIFAR-100]{
			\label{fig:app-cifar100-test}
			\includegraphics[width=0.235\linewidth]{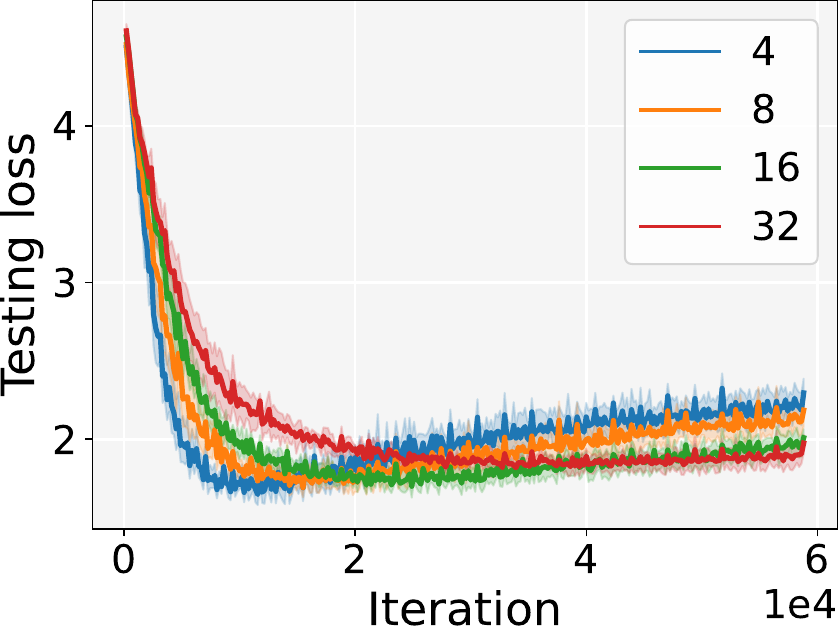}}
		\caption{Experimental result of delayed SGD on three non-convex problems, namely FC+MNIST (Figure (a)), ResNet-18+CIFAR-10 (Figure (b)), and ResNet-18+CIFAR-100 (illustrated in Figure \ref{fig:gen_resnet}). Figures (c) and (d) depict the training and testing loss of CIFAR-100. The variance in the plots is due to selecting different random seeds in multiple trials.}
		\label{fig:gen-nonconvex}
	\end{figure*}
	
	\section{Experimental validation}
	\label{sec:exp}
	In this section, we conduct simulation experiments to validate our theoretical findings for the quadratic convex model \eqref{erm}, where the generalization error plotted is the difference between the training loss and the testing loss. We used four classical datasets from the LIBSVM database \cite{Chang2011}, namely \texttt{rcv1} ($d=47,236$, $n=20,242$), \texttt{gisette} ($d=5,000$, $n=6,000$), \texttt{covtype} ($d=54$, $n=581,012$) and \texttt{ijcnn1} ($d=22$, $n=49,990$). Our simulation employed 16 distributed workers with a single local batch size of 16. We completed the experiments using the PyTorch framework on NVIDIA RTX 3090 GPUs, and repeated all experiments five times with different random seeds to eliminate interference.
	
	As discussed in the literature \cite{hardt2016train, regatti2019distributed}, the learning rate has a significant impact on generalization. More specifically, reducing the learning rate can make the algorithm more stable and thus reduce the generalization error. In the experiments, we used the same learning rate for training with different delays to clearly demonstrate the relationship between asynchronous delay and generalization error, despite our theory requiring a delay-dependent learning rate that satisfies $\eta<1/20\mu(\tau+1)$. Additionally, even though our theory has a learning rate requirement, we selected the optimal learning rate by grid search to minimize the empirical risk in our experiments. More specifically, for fixed delays, the learning rates for the four datasets \texttt{rcv1}, \texttt{gisette}, \texttt{covtype} and \texttt{ijcnn1}, are set to fixed values of $0.01$, $0.00005$, $0.0001$, and $0.01$, whereas for random delays, the corresponding learning rates are set to $0.05$, $0.00001$, $0.0001$, and $0.005$. The relatively small learning rates on the \texttt{covtype} and \texttt{gisette} datasets were determined by the characteristics of the datasets, and increasing the learning rates would make the training fail.

	Figure \ref{fig:gen_fixed} illustrate the generalization errors with fixed $4,8,16,32$ delays, while Figure \ref{fig:gen_random} present the results for random delays, with the legend representing the maximum delay $\overline{\tau}$. As shown in the figures, the generalization error grows with the iterations, but increasing the asynchronous delay makes this error decrease, which is consistent with our theoretical findings. Additionally, we showcase the training and testing loss curves of the \texttt{ijcnn1} data set in Figures \ref{fig:loss}, indicating that the asynchronous delay mitigates the overfitting phenomenon when the model converges, thereby leading to a decreased generalization error.

	We also conducted experiments on the non-convex problems (Figure \ref{fig:gen-nonconvex}) to further support the arguments in Figure \ref{fig:gen_resnet}. We utilized a three-layer fully connected (FC) neural network and ResNet-18 \cite{he2016deep} to classify the popular MNIST, CIFAR-10, and CIFAR-100 datasets \cite{lecun1998gradient, krizhevsky2009learning} with fixed learning rates of $0.01$, $0.01$, and $0.1$, respectively. The results of these non-convex experiments match well with our theoretical findings for the quadratic problems, suggesting that our theoretical results can be extended to non-convex optimization in some way.

	\section{Conclusion and future work}
	\label{sec:con}
	In this paper, we study the generalization performance of the delayed stochastic gradient descent algorithm. For the quadratic problem, we use the generating functions to derive the average stability of SGD with fixed and random delays. Building on the algorithmic stability, we provide sharper generalization error bounds for delayed SGD in the convex and strongly convex cases. Our theoretical results indicate that asynchronous delays make SGD more stable and hence reduce the generalization error. The corresponding experiments corroborate our theoretical findings.
	
	In practice, the delayed gradient method exhibits similar generalization performance in non-convex problems, as illustrated in Figures \ref{fig:gen_resnet} and \ref{fig:gen-nonconvex}. Therefore, our current findings may be extended to the widespread non-convex applications in the future. Additionally, SGD is proven to be robust to arbitrary delays in the optimization theory \cite{cohen2021asynchronous, mishchenko2022asynchronous}. Hence an interesting direction for future research is to investigate the generalization performance of SGD with arbitrary delays.

	\ifCLASSOPTIONcaptionsoff
	\newpage
	\fi

	
	
	\bibliographystyle{IEEEtran}
	\bibliography{ref}

\begin{thebibliography}{10}
\providecommand{\url}[1]{#1}
\csname url@samestyle\endcsname
\providecommand{\newblock}{\relax}
\providecommand{\bibinfo}[2]{#2}
\providecommand{\BIBentrySTDinterwordspacing}{\spaceskip=0pt\relax}
\providecommand{\BIBentryALTinterwordstretchfactor}{4}
\providecommand{\BIBentryALTinterwordspacing}{\spaceskip=\fontdimen2\font plus
\BIBentryALTinterwordstretchfactor\fontdimen3\font minus
  \fontdimen4\font\relax}
\providecommand{\BIBforeignlanguage}[2]{{%
\expandafter\ifx\csname l@#1\endcsname\relax
\typeout{** WARNING: IEEEtran.bst: No hyphenation pattern has been}%
\typeout{** loaded for the language `#1'. Using the pattern for}%
\typeout{** the default language instead.}%
\else
\language=\csname l@#1\endcsname
\fi
#2}}
\providecommand{\BIBdecl}{\relax}
\BIBdecl

\bibitem{robbins1951stochastic}
H.~Robbins and S.~Monro, ``A stochastic approximation method,'' \emph{The
  Annals of Mathematical Statistics}, vol.~22, no.~3, pp. 400--407, 1951.

\bibitem{bottou2018optimization}
L.~Bottou, F.~E. Curtis, and J.~Nocedal, ``Optimization methods for large-scale
  machine learning,'' \emph{SIAM review}, vol.~60, no.~2, pp. 223--311, 2018.

\bibitem{zhang2017understanding}
C.~Zhang, S.~Bengio, M.~Hardt, B.~Recht, and O.~Vinyals, ``Understanding deep
  learning requires rethinking generalization,'' in \emph{International
  Conference on Learning Representations}, 2017.

\bibitem{du2018gradient}
S.~S. Du, X.~Zhai, B.~Poczos, and A.~Singh, ``Gradient descent provably
  optimizes over-parameterized neural networks,'' in \emph{International
  Conference on Learning Representations}, 2019.

\bibitem{hardt2016train}
M.~Hardt, B.~Recht, and Y.~Singer, ``Train faster, generalize better:
  {S}tability of stochastic gradient descent,'' in \emph{International
  Conference on Machine Learning}, M.~F. Balcan and K.~Q. Weinberger, Eds.,
  vol.~48.\hskip 1em plus 0.5em minus 0.4em\relax PMLR, 2016, pp. 1225--1234.

\bibitem{kuzborskij2018data}
I.~Kuzborskij and C.~Lampert, ``Data-dependent stability of stochastic gradient
  descent,'' in \emph{International Conference on Machine Learning}, J.~Dy and
  A.~Krause, Eds., vol.~80.\hskip 1em plus 0.5em minus 0.4em\relax PMLR, 2018,
  pp. 2815--2824.

\bibitem{lei2020fine}
Y.~Lei and Y.~Ying, ``Fine-grained analysis of stability and generalization for
  stochastic gradient descent,'' in \emph{International Conference on Machine
  Learning}, H.~D. III and A.~Singh, Eds., vol. 119.\hskip 1em plus 0.5em minus
  0.4em\relax PMLR, 2020, pp. 5809--5819.

\bibitem{bassily2020stability}
R.~Bassily, V.~Feldman, C.~Guzm{\'a}n, and K.~Talwar, ``Stability of stochastic
  gradient descent on nonsmooth convex losses,'' in \emph{Advances in Neural
  Information Processing Systems}, H.~Larochelle, M.~Ranzato, R.~Hadsell,
  M.~Balcan, and H.~Lin, Eds., vol.~33.\hskip 1em plus 0.5em minus 0.4em\relax
  Curran Associates, Inc., 2020, pp. 4381--4391.

\bibitem{zhang2022stability}
Y.~Zhang, W.~Zhang, S.~Bald, V.~Pingali, C.~Chen, and M.~Goswami, ``Stability
  of {SGD}: {T}ightness analysis and improved bounds,'' in \emph{Uncertainty in
  Artificial Intelligence}, J.~Cussens and K.~Zhang, Eds., vol. 180.\hskip 1em
  plus 0.5em minus 0.4em\relax PMLR, 2022, pp. 2364--2373.

\bibitem{zhou2022understanding}
Y.~Zhou, Y.~Liang, and H.~Zhang, ``Understanding generalization error of {SGD}
  in nonconvex optimization,'' \emph{Machine Learning}, vol. 111, no.~1, pp.
  345--375, 2022.

\bibitem{deng2009imagenet}
J.~Deng, W.~Dong, R.~Socher, L.-J. Li, K.~Li, and L.~Fei-Fei, ``Imagenet: {A}
  large-scale hierarchical image database,'' in \emph{2009 IEEE Conference on
  Computer Vision and Pattern Recognition}.\hskip 1em plus 0.5em minus
  0.4em\relax IEEE, 2009, pp. 248--255.

\bibitem{dean2012large}
J.~Dean, G.~Corrado, R.~Monga, K.~Chen, M.~Devin, M.~Mao, M.~Ranzato,
  A.~Senior, P.~Tucker, K.~Yang \emph{et~al.}, ``Large scale distributed deep
  networks,'' in \emph{Advances in Neural Information Processing Systems},
  F.~Pereira, C.~Burges, L.~Bottou, and K.~Weinberger, Eds., vol.~25.\hskip 1em
  plus 0.5em minus 0.4em\relax Curran Associates, Inc., 2012, pp. 1223--1231.

\bibitem{brown2020language}
T.~Brown, B.~Mann, N.~Ryder, M.~Subbiah, J.~D. Kaplan, P.~Dhariwal,
  A.~Neelakantan, P.~Shyam, G.~Sastry, A.~Askell \emph{et~al.}, ``Language
  models are few-shot learners,'' in \emph{Advances in Neural Information
  Processing Systems}, H.~Larochelle, M.~Ranzato, R.~Hadsell, M.~Balcan, and
  H.~Lin, Eds., vol.~33.\hskip 1em plus 0.5em minus 0.4em\relax Curran
  Associates, Inc., 2020, pp. 1877--1901.

\bibitem{zinkevich2010parallelized}
M.~Zinkevich, M.~Weimer, L.~Li, and A.~Smola, ``Parallelized stochastic
  gradient descent,'' in \emph{Advances in Neural Information Processing
  Systems}, J.~Lafferty, C.~Williams, J.~Shawe-Taylor, R.~Zemel, and
  A.~Culotta, Eds., vol.~23.\hskip 1em plus 0.5em minus 0.4em\relax Curran
  Associates, Inc., 2010.

\bibitem{li2014scaling}
M.~Li, D.~G. Andersen, J.~W. Park, A.~J. Smola, A.~Ahmed, V.~Josifovski,
  J.~Long, E.~J. Shekita, and B.-Y. Su, ``Scaling distributed machine learning
  with the parameter server,'' in \emph{11th USENIX Symposium on Operating
  Systems Design and Implementation (OSDI 14)}.\hskip 1em plus 0.5em minus
  0.4em\relax USENIX Association, 2014, pp. 583--598.

\bibitem{assran2020advances}
M.~Assran, A.~Aytekin, H.~R. Feyzmahdavian, M.~Johansson, and M.~G. Rabbat,
  ``Advances in asynchronous parallel and distributed optimization,''
  \emph{Proceedings of the IEEE}, vol. 108, no.~11, pp. 2013--2031, 2020.

\bibitem{nedic2001distributed}
A.~Nedi{\'c}, D.~P. Bertsekas, and V.~S. Borkar, ``Distributed asynchronous
  incremental subgradient methods,'' \emph{Studies in Computational
  Mathematics}, vol.~8, no.~C, pp. 381--407, 2001.

\bibitem{agarwal2011distributed}
A.~Agarwal and J.~C. Duchi, ``Distributed delayed stochastic optimization,'' in
  \emph{Advances in Neural Information Processing Systems}, J.~Shawe-Taylor,
  R.~Zemel, P.~Bartlett, F.~Pereira, and K.~Weinberger, Eds., vol.~24.\hskip
  1em plus 0.5em minus 0.4em\relax Curran Associates, Inc., 2011.

\bibitem{lian2015asynchronous}
X.~Lian, Y.~Huang, Y.~Li, and J.~Liu, ``Asynchronous parallel stochastic
  gradient for nonconvex optimization,'' in \emph{Advances in Neural
  Information Processing Systems}, C.~Cortes, N.~Lawrence, D.~Lee, M.~Sugiyama,
  and R.~Garnett, Eds., vol.~28.\hskip 1em plus 0.5em minus 0.4em\relax Curran
  Associates, Inc., 2015.

\bibitem{zhou2018distributed}
Z.~Zhou, P.~Mertikopoulos, N.~Bambos, P.~Glynn, Y.~Ye, L.-J. Li, and
  L.~Fei-Fei, ``Distributed asynchronous optimization with unbounded delays:
  {H}ow slow can you go?'' in \emph{International Conference on Machine
  Learning}, J.~Dy and A.~Krause, Eds., vol.~80.\hskip 1em plus 0.5em minus
  0.4em\relax PMLR, 2018, pp. 5970--5979.

\bibitem{arjevani2020tight}
Y.~Arjevani, O.~Shamir, and N.~Srebro, ``A tight convergence analysis for
  stochastic gradient descent with delayed updates,'' in \emph{Proceedings of
  the 31st International Conference on Algorithmic Learning Theory},
  A.~Kontorovich and G.~Neu, Eds., vol. 117.\hskip 1em plus 0.5em minus
  0.4em\relax PMLR, 2020, pp. 111--132.

\bibitem{stich2020error}
S.~U. Stich and S.~P. Karimireddy, ``The error-feedback framework: {B}etter
  rates for {SGD} with delayed gradients and compressed updates,''
  \emph{Journal of Machine Learning Research}, vol.~21, no. 237, pp. 1--36,
  2020.

\bibitem{cohen2021asynchronous}
A.~Cohen, A.~Daniely, Y.~Drori, T.~Koren, and M.~Schain, ``Asynchronous
  stochastic optimization robust to arbitrary delays,'' in \emph{Advances in
  Neural Information Processing Systems}, M.~Ranzato, A.~Beygelzimer,
  Y.~Dauphin, P.~Liang, and J.~W. Vaughan, Eds., vol.~34.\hskip 1em plus 0.5em
  minus 0.4em\relax Curran Associates, Inc., 2021, pp. 9024--9035.

\bibitem{mishchenko2022asynchronous}
K.~Mishchenko, F.~Bach, M.~Even, and B.~E. Woodworth, ``Asynchronous {SGD}
  beats minibatch {SGD} under arbitrary delays,'' in \emph{Advances in Neural
  Information Processing Systems}, S.~Koyejo, S.~Mohamed, A.~Agarwal,
  D.~Belgrave, K.~Cho, and A.~Oh, Eds., vol.~35.\hskip 1em plus 0.5em minus
  0.4em\relax Curran Associates, Inc., 2022, pp. 420--433.

\bibitem{regatti2019distributed}
J.~Regatti, G.~Tendolkar, Y.~Zhou, A.~Gupta, and Y.~Liang, ``Distributed {SGD}
  generalizes well under asynchrony,'' in \emph{2019 57th Annual Allerton
  Conference on Communication, Control, and Computing (Allerton)}.\hskip 1em
  plus 0.5em minus 0.4em\relax IEEE, 2019, pp. 863--870.

\bibitem{li2018visualizing}
H.~Li, Z.~Xu, G.~Taylor, C.~Studer, and T.~Goldstein, ``Visualizing the loss
  landscape of neural nets,'' in \emph{Advances in Neural Information
  Processing Systems}, S.~Bengio, H.~Wallach, H.~Larochelle, K.~Grauman,
  N.~Cesa-Bianchi, and R.~Garnett, Eds., vol.~31.\hskip 1em plus 0.5em minus
  0.4em\relax Curran Associates, Inc., 2018.

\bibitem{jacot2018neural}
A.~Jacot, F.~Gabriel, and C.~Hongler, ``Neural tangent kernel: {C}onvergence
  and generalization in neural networks,'' in \emph{Advances in Neural
  Information Processing Systems}, S.~Bengio, H.~Wallach, H.~Larochelle,
  K.~Grauman, N.~Cesa-Bianchi, and R.~Garnett, Eds., vol.~31.\hskip 1em plus
  0.5em minus 0.4em\relax Curran Associates, Inc., 2018.

\bibitem{ma2018power}
S.~Ma, R.~Bassily, and M.~Belkin, ``The power of interpolation: {U}nderstanding
  the effectiveness of {SGD} in modern over-parametrized learning,'' in
  \emph{International Conference on Machine Learning}, J.~Dy and A.~Krause,
  Eds., vol.~80.\hskip 1em plus 0.5em minus 0.4em\relax PMLR, 2018, pp.
  3325--3334.

\bibitem{he2019control}
F.~He, T.~Liu, and D.~Tao, ``Control batch size and learning rate to generalize
  well: {T}heoretical and empirical evidence,'' in \emph{Advances in Neural
  Information Processing Systems}, H.~Wallach, H.~Larochelle, A.~Beygelzimer,
  F.~d\textquotesingle Alch\'{e}-Buc, E.~Fox, and R.~Garnett, Eds.,
  vol.~32.\hskip 1em plus 0.5em minus 0.4em\relax Curran Associates, Inc.,
  2019.

\bibitem{zou2021benign}
D.~Zou, J.~Wu, V.~Braverman, Q.~Gu, and S.~Kakade, ``Benign overfitting of
  constant-stepsize {SGD} for linear regression,'' in \emph{Proceedings of
  Thirty Fourth Conference on Learning Theory}, M.~Belkin and S.~Kpotufe, Eds.,
  vol. 134.\hskip 1em plus 0.5em minus 0.4em\relax PMLR, 2021, pp. 4633--4635.

\bibitem{tsitsiklis1986distributed}
J.~Tsitsiklis, D.~Bertsekas, and M.~Athans, ``Distributed asynchronous
  deterministic and stochastic gradient optimization algorithms,'' \emph{IEEE
  Transactions on Automatic Control}, vol.~31, no.~9, pp. 803--812, 1986.

\bibitem{recht2011hogwild}
B.~Recht, C.~Re, S.~Wright, and F.~Niu, ``Hogwild!: {A} lock-free approach to
  parallelizing stochastic gradient descent,'' in \emph{Advances in Neural
  Information Processing Systems}, J.~Shawe-Taylor, R.~Zemel, P.~Bartlett,
  F.~Pereira, and K.~Weinberger, Eds., vol.~24.\hskip 1em plus 0.5em minus
  0.4em\relax Curran Associates, Inc., 2011.

\bibitem{mania2015perturbed}
H.~Mania, X.~Pan, D.~Papailiopoulos, B.~Recht, K.~Ramchandran, and M.~I.
  Jordan, ``Perturbed iterate analysis for asynchronous stochastic
  optimization,'' \emph{SIAM Journal on Optimization}, vol.~27, no.~4, pp.
  2202--2229, 2017.

\bibitem{sun2017asynchronous}
T.~Sun, R.~Hannah, and W.~Yin, ``Asynchronous coordinate descent under more
  realistic assumptions,'' in \emph{Advances in Neural Information Processing
  Systems}, I.~Guyon, U.~V. Luxburg, S.~Bengio, H.~Wallach, R.~Fergus,
  S.~Vishwanathan, and R.~Garnett, Eds., vol.~30.\hskip 1em plus 0.5em minus
  0.4em\relax Curran Associates, Inc., 2017.

\bibitem{aviv2021asynchronous}
R.~Z. Aviv, I.~Hakimi, A.~Schuster, and K.~Y. Levy, ``Asynchronous distributed
  learning: {A}dapting to gradient delays without prior knowledge,'' in
  \emph{International Conference on Machine Learning}, M.~Meila and T.~Zhang,
  Eds., vol. 139.\hskip 1em plus 0.5em minus 0.4em\relax PMLR, 2021, pp.
  436--445.

\bibitem{zheng2017asynchronous}
S.~Zheng, Q.~Meng, T.~Wang, W.~Chen, N.~Yu, Z.-M. Ma, and T.-Y. Liu,
  ``Asynchronous stochastic gradient descent with delay compensation,'' in
  \emph{International Conference on Machine Learning}, D.~Precup and Y.~W. Teh,
  Eds., vol.~70.\hskip 1em plus 0.5em minus 0.4em\relax PMLR, 2017, pp.
  4120--4129.

\bibitem{sra2016adadelay}
S.~Sra, A.~W. Yu, M.~Li, and A.~Smola, ``Adadelay: Delay adaptive distributed
  stochastic optimization,'' in \emph{Artificial Intelligence and Statistics},
  A.~Gretton and C.~C. Robert, Eds., vol.~51.\hskip 1em plus 0.5em minus
  0.4em\relax PMLR, 2016, pp. 957--965.

\bibitem{zhang2016staleness}
W.~Zhang, S.~Gupta, X.~Lian, and J.~Liu, ``Staleness-aware async-{SGD} for
  distributed deep learning,'' in \emph{Proceedings of the Twenty-Fifth
  International Joint Conference on Artificial Intelligence}, S.~Kambhampati,
  Ed.\hskip 1em plus 0.5em minus 0.4em\relax {IJCAI/AAAI} Press, 2016, pp.
  2350--2356.

\bibitem{ren2020delay}
Z.~Ren, Z.~Zhou, L.~Qiu, A.~Deshpande, and J.~Kalagnanam, ``Delay-adaptive
  distributed stochastic optimization,'' in \emph{Proceedings of the AAAI
  Conference on Artificial Intelligence}, vol.~34, 2020, pp. 5503--5510.

\bibitem{backstrom2022asap}
K.~B{\"a}ckstr{\"o}m, M.~Papatriantafilou, and P.~Tsigas, ``{ASAP}.{SGD}:
  {I}nstance-based adaptiveness to staleness in asynchronous {SGD},'' in
  \emph{Proceedings of the 39th International Conference on Machine Learning},
  K.~Chaudhuri, S.~Jegelka, L.~Song, C.~Szepesvari, G.~Niu, and S.~Sabato,
  Eds., vol. 162.\hskip 1em plus 0.5em minus 0.4em\relax PMLR, 2022, pp.
  1261--1276.

\bibitem{mnih2016asynchronous}
V.~Mnih, A.~P. Badia, M.~Mirza, A.~Graves, T.~Lillicrap, T.~Harley, D.~Silver,
  and K.~Kavukcuoglu, ``Asynchronous methods for deep reinforcement learning,''
  in \emph{International conference on machine learning}, M.~F. Balcan and
  K.~Q. Weinberger, Eds., vol.~48.\hskip 1em plus 0.5em minus 0.4em\relax PMLR,
  2016, pp. 1928--1937.

\bibitem{koloskova2022sharper}
A.~Koloskova, S.~U. Stich, and M.~Jaggi, ``Sharper convergence guarantees for
  asynchronous {SGD} for distributed and federated learning,'' in
  \emph{Advances in Neural Information Processing Systems}, S.~Koyejo,
  S.~Mohamed, A.~Agarwal, D.~Belgrave, K.~Cho, and A.~Oh, Eds., vol.~35.\hskip
  1em plus 0.5em minus 0.4em\relax Curran Associates, Inc., 2022, pp.
  17\,202--17\,215.

\bibitem{hsieh2022multi}
Y.-G. Hsieh, F.~Iutzeler, J.~Malick, and P.~Mertikopoulos, ``Multi-agent online
  optimization with delays: Asynchronicity, adaptivity, and optimism,''
  \emph{Journal of Machine Learning Research}, vol.~23, no.~78, pp. 1--49,
  2022.

\bibitem{rogers1978finite}
W.~H. Rogers and T.~J. Wagner, ``A finite sample distribution-free performance
  bound for local discrimination rules,'' \emph{The Annals of Statistics},
  vol.~6, no.~3, pp. 506--514, 1978.

\bibitem{devroye1979distribution1}
L.~Devroye and T.~Wagner, ``Distribution-free inequalities for the deleted and
  holdout error estimates,'' \emph{IEEE Transactions on Information Theory},
  vol.~25, no.~2, pp. 202--207, 1979.

\bibitem{devroye1979distribution2}
------, ``Distribution-free performance bounds with the resubstitution error
  estimate (corresp.),'' \emph{IEEE Transactions on Information Theory},
  vol.~25, no.~2, pp. 208--210, 1979.

\bibitem{bousquet2002stability}
O.~Bousquet and A.~Elisseeff, ``Stability and generalization,'' \emph{Journal
  of Machine Learning Research}, vol.~2, pp. 499--526, 2002.

\bibitem{elisseeff2005stability}
A.~Elisseeff, T.~Evgeniou, M.~Pontil, and L.~P. Kaelbing, ``Stability of
  randomized learning algorithms.'' \emph{Journal of Machine Learning
  Research}, vol.~6, no.~3, pp. 55--79, 2005.

\bibitem{mukherjee2006learning}
S.~Mukherjee, P.~Niyogi, T.~Poggio, and R.~Rifkin, ``Learning theory: stability
  is sufficient for generalization and necessary and sufficient for consistency
  of empirical risk minimization,'' \emph{Advances in Computational
  Mathematics}, vol.~25, no. 1-3, pp. 161--193, 2006.

\bibitem{shalev2010learnability}
S.~Shalev-Shwartz, O.~Shamir, N.~Srebro, and K.~Sridharan, ``Learnability,
  stability and uniform convergence,'' \emph{Journal of Machine Learning
  Research}, vol.~11, no.~90, pp. 2635--2670, 2010.

\bibitem{charles2018stability}
Z.~Charles and D.~Papailiopoulos, ``Stability and generalization of learning
  algorithms that converge to global optima,'' in \emph{International
  Conference on Machine Learning}, J.~Dy and A.~Krause, Eds., vol.~80.\hskip
  1em plus 0.5em minus 0.4em\relax PMLR, 2018, pp. 745--754.

\bibitem{feldman2018generalization}
V.~Feldman and J.~Vondrak, ``Generalization bounds for uniformly stable
  algorithms,'' in \emph{Advances in Neural Information Processing Systems},
  S.~Bengio, H.~Wallach, H.~Larochelle, K.~Grauman, N.~Cesa-Bianchi, and
  R.~Garnett, Eds., vol.~31.\hskip 1em plus 0.5em minus 0.4em\relax Curran
  Associates, Inc., 2018.

\bibitem{feldman2019high}
------, ``High probability generalization bounds for uniformly stable
  algorithms with nearly optimal rate,'' in \emph{Conference on Learning
  Theory}, A.~Beygelzimer and D.~Hsu, Eds., vol.~99.\hskip 1em plus 0.5em minus
  0.4em\relax PMLR, 2019, pp. 1270--1279.

\bibitem{bousquet2020sharper}
O.~Bousquet, Y.~Klochkov, and N.~Zhivotovskiy, ``Sharper bounds for uniformly
  stable algorithms,'' in \emph{Conference on Learning Theory}, J.~Abernethy
  and S.~Agarwal, Eds., vol. 125.\hskip 1em plus 0.5em minus 0.4em\relax PMLR,
  2020, pp. 610--626.

\bibitem{chandramoorthy2022on}
N.~Chandramoorthy, A.~Loukas, K.~Gatmiry, and S.~Jegelka, ``On the
  generalization of learning algorithms that do not converge,'' in
  \emph{Advances in Neural Information Processing Systems}, S.~Koyejo,
  S.~Mohamed, A.~Agarwal, D.~Belgrave, K.~Cho, and A.~Oh, Eds., vol.~35.\hskip
  1em plus 0.5em minus 0.4em\relax Curran Associates, Inc., 2022, pp.
  34\,241--34\,257.

\bibitem{richards2021stability}
D.~Richards and I.~Kuzborskij, ``Stability \& generalisation of gradient
  descent for shallow neural networks without the neural tangent kernel,'' in
  \emph{Advances in Neural Information Processing Systems}, M.~Ranzato,
  A.~Beygelzimer, Y.~Dauphin, P.~Liang, and J.~W. Vaughan, Eds., vol.~34.\hskip
  1em plus 0.5em minus 0.4em\relax Curran Associates, Inc., 2021, pp.
  8609--8621.

\bibitem{lei2022stability}
Y.~Lei, R.~Jin, and Y.~Ying, ``Stability and generalization analysis of
  gradient methods for shallow neural networks,'' in \emph{Advances in Neural
  Information Processing Systems}, S.~Koyejo, S.~Mohamed, A.~Agarwal,
  D.~Belgrave, K.~Cho, and A.~Oh, Eds., vol.~35.\hskip 1em plus 0.5em minus
  0.4em\relax Curran Associates, Inc., 2022, pp. 38\,557--38\,570.

\bibitem{mou2018generalization}
W.~Mou, L.~Wang, X.~Zhai, and K.~Zheng, ``Generalization bounds of {SGLD} for
  non-convex learning: {T}wo theoretical viewpoints,'' in \emph{Conference on
  Learning Theory}, S.~Bubeck, V.~Perchet, and P.~Rigollet, Eds.,
  vol.~75.\hskip 1em plus 0.5em minus 0.4em\relax PMLR, 2018, pp. 605--638.

\bibitem{banerjee2022stability}
A.~Banerjee, T.~Chen, X.~Li, and Y.~Zhou, ``Stability based generalization
  bounds for exponential family langevin dynamics,'' in \emph{International
  Conference on Machine Learning}, K.~Chaudhuri, S.~Jegelka, L.~Song,
  C.~Szepesvari, G.~Niu, and S.~Sabato, Eds., vol. 162.\hskip 1em plus 0.5em
  minus 0.4em\relax PMLR, 2022, pp. 1412--1449.

\bibitem{lei2020sharper}
Y.~Lei, A.~Ledent, and M.~Kloft, ``Sharper generalization bounds for pairwise
  learning,'' in \emph{Advances in Neural Information Processing Systems},
  H.~Larochelle, M.~Ranzato, R.~Hadsell, M.~Balcan, and H.~Lin, Eds.,
  vol.~33.\hskip 1em plus 0.5em minus 0.4em\relax Curran Associates, Inc.,
  2020, pp. 21\,236--21\,246.

\bibitem{yang2021simple}
Z.~Yang, Y.~Lei, P.~Wang, T.~Yang, and Y.~Ying, ``Simple stochastic and online
  gradient descent algorithms for pairwise learning,'' in \emph{Advances in
  Neural Information Processing Systems}, M.~Ranzato, A.~Beygelzimer,
  Y.~Dauphin, P.~Liang, and J.~W. Vaughan, Eds., vol.~34.\hskip 1em plus 0.5em
  minus 0.4em\relax Curran Associates, Inc., 2021, pp. 20\,160--20\,171.

\bibitem{farid2021generalization}
A.~Farid and A.~Majumdar, ``Generalization bounds for meta-learning via
  {PAC}-{B}ayes and uniform stability,'' in \emph{Advances in Neural
  Information Processing Systems}, M.~Ranzato, A.~Beygelzimer, Y.~Dauphin,
  P.~Liang, and J.~W. Vaughan, Eds., vol.~34.\hskip 1em plus 0.5em minus
  0.4em\relax Curran Associates, Inc., 2021, pp. 2173--2186.

\bibitem{guan2022finegrained}
J.~Guan, Y.~Liu, and Z.~Lu, ``Fine-grained analysis of stability and
  generalization for modern meta learning algorithms,'' in \emph{Advances in
  Neural Information Processing Systems}, S.~Koyejo, S.~Mohamed, A.~Agarwal,
  D.~Belgrave, K.~Cho, and A.~Oh, Eds., vol.~35.\hskip 1em plus 0.5em minus
  0.4em\relax Curran Associates, Inc., 2022, pp. 18\,487--18\,500.

\bibitem{farnia2021train}
F.~Farnia and A.~Ozdaglar, ``Train simultaneously, generalize better:
  {S}tability of gradient-based minimax learners,'' in \emph{International
  Conference on Machine Learning}, M.~Meila and T.~Zhang, Eds., vol. 139.\hskip
  1em plus 0.5em minus 0.4em\relax PMLR, 2021, pp. 3174--3185.

\bibitem{lei2021stability}
Y.~Lei, Z.~Yang, T.~Yang, and Y.~Ying, ``Stability and generalization of
  stochastic gradient methods for minimax problems,'' in \emph{International
  Conference on Machine Learning}, M.~Meila and T.~Zhang, Eds., vol. 139.\hskip
  1em plus 0.5em minus 0.4em\relax PMLR, 2021, pp. 6175--6186.

\bibitem{xing2021algorithmic}
Y.~Xing, Q.~Song, and G.~Cheng, ``On the algorithmic stability of adversarial
  training,'' in \emph{Advances in Neural Information Processing Systems},
  M.~Ranzato, A.~Beygelzimer, Y.~Dauphin, P.~Liang, and J.~W. Vaughan, Eds.,
  vol.~34.\hskip 1em plus 0.5em minus 0.4em\relax Curran Associates, Inc.,
  2021, pp. 26\,523--26\,535.

\bibitem{NEURIPS2022_f9b8853e}
A.~Ozdaglar, S.~Pattathil, J.~Zhang, and K.~Zhang, ``What is a good metric to
  study generalization of minimax learners?'' in \emph{Advances in Neural
  Information Processing Systems}, S.~Koyejo, S.~Mohamed, A.~Agarwal,
  D.~Belgrave, K.~Cho, and A.~Oh, Eds., vol.~35.\hskip 1em plus 0.5em minus
  0.4em\relax Curran Associates, Inc., 2022, pp. 38\,190--38\,203.

\bibitem{xiao2022stability}
J.~Xiao, Y.~Fan, R.~Sun, J.~Wang, and Z.-Q. Luo, ``Stability analysis and
  generalization bounds of adversarial training,'' in \emph{Advances in Neural
  Information Processing Systems}, S.~Koyejo, S.~Mohamed, A.~Agarwal,
  D.~Belgrave, K.~Cho, and A.~Oh, Eds., vol.~35.\hskip 1em plus 0.5em minus
  0.4em\relax Curran Associates, Inc., 2022, pp. 15\,446--15\,459.

\bibitem{wu2019stability}
X.~Wu, J.~Zhang, and F.-Y. Wang, ``Stability-based generalization analysis of
  distributed learning algorithms for big data,'' \emph{IEEE Transactions on
  Neural Networks and Learning Systems}, vol.~31, no.~3, pp. 801--812, 2019.

\bibitem{sun2021stability}
T.~Sun, D.~Li, and B.~Wang, ``Stability and generalization of decentralized
  stochastic gradient descent,'' in \emph{Proceedings of the AAAI Conference on
  Artificial Intelligence}, vol.~35, 2021, pp. 9756--9764.

\bibitem{deng2023stability}
X.~Deng, T.~Sun, S.~Li, and D.~Li, ``Stability-based generalization analysis of
  the asynchronous decentralized {SGD},'' in \emph{Proceedings of the AAAI
  Conference on Artificial Intelligence}, vol.~37, no.~6, 2023, pp. 7340--7348.

\bibitem{zhu2022topology}
T.~Zhu, F.~He, L.~Zhang, Z.~Niu, M.~Song, and D.~Tao, ``Topology-aware
  generalization of decentralized {SGD},'' in \emph{International Conference on
  Machine Learning}, K.~Chaudhuri, S.~Jegelka, L.~Song, C.~Szepesvari, G.~Niu,
  and S.~Sabato, Eds., vol. 162.\hskip 1em plus 0.5em minus 0.4em\relax PMLR,
  2022, pp. 27\,479--27\,503.

\bibitem{WILF199027}
H.~S. Wilf, ``Chapter 2 - series,'' in \emph{Generatingfunctionology}, H.~S.
  Wilf, Ed.\hskip 1em plus 0.5em minus 0.4em\relax Academic Press, 1990, pp.
  27--63.

\bibitem{stanley_fomin_1999}
R.~P. Stanley and S.~Fomin, \emph{Enumerative Combinatorics}, ser. Cambridge
  Studies in Advanced Mathematics.\hskip 1em plus 0.5em minus 0.4em\relax
  Cambridge University Press, 1999, vol.~2.

\bibitem{flajolet_sedgewick_2009}
P.~Flajolet and R.~Sedgewick, \emph{Combinatorial structures and ordinary
  generating functions}.\hskip 1em plus 0.5em minus 0.4em\relax Cambridge
  University Press, 2009, p. 15–94.

\bibitem{NEURIPS2019_0d1a9651}
J.~Lee, L.~Xiao, S.~Schoenholz, Y.~Bahri, R.~Novak, J.~Sohl-Dickstein, and
  J.~Pennington, ``Wide neural networks of any depth evolve as linear models
  under gradient descent,'' in \emph{Advances in Neural Information Processing
  Systems}, H.~Wallach, H.~Larochelle, A.~Beygelzimer, F.~d\textquotesingle
  Alch\'{e}-Buc, E.~Fox, and R.~Garnett, Eds., vol.~32.\hskip 1em plus 0.5em
  minus 0.4em\relax Curran Associates, Inc., 2019.

\bibitem{Chang2011}
C.-C. Chang and C.-J. Lin, ``{LIBSVM}: {A} library for support vector
  machines,'' \emph{ACM Transactions on Intelligent Systems and Technology},
  vol.~2, pp. 27:1--27:27, 2011.

\bibitem{he2016deep}
K.~He, X.~Zhang, S.~Ren, and J.~Sun, ``Deep residual learning for image
  recognition,'' in \emph{IEEE Conference on Computer Vision and Pattern
  Recognition}, 2016, pp. 770--778.

\bibitem{lecun1998gradient}
Y.~LeCun, L.~Bottou, Y.~Bengio, and P.~Haffner, ``Gradient-based learning
  applied to document recognition,'' \emph{Proceedings of the IEEE}, vol.~86,
  no.~11, pp. 2278--2324, 1998.

\bibitem{krizhevsky2009learning}
A.~Krizhevsky, G.~Hinton \emph{et~al.}, ``Learning multiple layers of features
  from tiny images,'' pp. 32--33, 2009.

\end{thebibliography}

\onecolumn
\appendices
\begin{center}
	{\Large \bf Supplementary materials for \vspace{1.5ex}\\
		\textit{\fontsize{11.5pt}{\baselineskip}\selectfont Towards Understanding the Generalizability of Delayed Stochastic Gradient Descent}}
\end{center}
\bigskip


\section{Proof of Lemma \ref{lem:1}}
The proof is based on an argument in Lemma 11 of \cite{shalev2010learnability}. For any $i$, the data samples $\mathbf{z}_{i}$ and $\mathbf{z}_{i}'$ are both drawn i.i.d. from $\mathcal{D}$. We denote the sets
\begin{equation}
	\nonumber
	\mathcal{S}=\{\mathbf{z}_{1}, \ldots, \mathbf{z}_{n}\}, ~\mathcal{S}'=\{\mathbf{z}_{1}', \ldots, \mathbf{z}_{n}'\}, ~\text{and}~~ \mathcal{S}^{(i)}=\{\mathbf{z}_{1}, \ldots, \mathbf{z}_{i-1}, \mathbf{z}_{i}', \mathbf{z}_{i+1}, \ldots, \mathbf{z}_{n}\}.
\end{equation}
Due to the fact that
\begin{equation}
	\label{fact:1}
	\begin{split}
		F(\mathcal{A}(\mathcal{S}))&=\mathbb{E}_{\mathcal{S}'}[f(\mathcal{A}(\mathcal{S}); \mathbf{z}_{i}')]=\frac{1}{n}\sum_{i=1}^{n}\mathbb{E}_{\mathcal{S}'}[f(\mathcal{A}(\mathcal{S}); \mathbf{z}_{i}')],\\ \mathbb{E}_{\mathcal{S}}[f(\mathcal{A}(\mathcal{S}); \mathbf{z}_{i})]&=\mathbb{E}_{\mathcal{S}, \mathcal{S}'}[f(\mathcal{A}(\mathcal{S}^{(i)}); \mathbf{z}_{i}')].
	\end{split}
\end{equation}
Hence,
\begin{equation}
	\nonumber
	\begin{split}
		\epsilon_{\text{gen}}=\mathbb{E}_{\mathcal{S}, \mathcal{A}}\left[F(\mathcal{A}(\mathcal{S}))-F_{\mathcal{S}}(\mathcal{A}(\mathcal{S}))\right]&=\mathbb{E}_{\mathcal{S}, \mathcal{A}}\left[\mathbb{E}_{\mathcal{S}'}[f(\mathcal{A}(\mathcal{S}); \mathbf{z}_{i}')]-\frac{1}{n}\sum_{i=1}^{n}f(\mathcal{A}(\mathcal{S}); \mathbf{z}_{i})\right]\\
		&=\frac{1}{n}\sum_{i=1}^{n}\mathbb{E}_{\mathcal{S}, \mathcal{A}}\Big[\mathbb{E}_{\mathcal{S}'}[f(\mathcal{A}(\mathcal{S}); \mathbf{z}_{i}')]-f(\mathcal{A}(\mathcal{S}); \mathbf{z}_{i})\Big]\\
		&=\frac{1}{n}\sum_{i=1}^{n}\mathbb{E}_{\mathcal{S}, \mathcal{S}', \mathcal{A}}\left[f(\mathcal{A}(\mathcal{S});\mathbf{z}_{i}')-f(\mathcal{A}(\mathcal{S}^{(i)});\mathbf{z}_{i}')\right]\\
		&\leq \epsilon_{\text{stab}}.
	\end{split}
\end{equation}
$\hfill\blacksquare$ 

\section{Proof of Remark \ref{rmk:0}}
Since for any $i$, $\mathbf{z}_{i}$ and $\mathbf{z}_{i}'$ are both drawn i.i.d. from $\mathcal{D}$, we can see that $\mathcal{A}(\mathcal{S}^{(i)})$ is independent of $\mathbf{z}_{i}$. Hence we can obtain the following fact, which is also applied in the proof of \cite{lei2020fine} (Appendix B of \cite{lei2020fine})
\begin{equation}
	\label{fact:2}
	\mathbb{E}_{\mathcal{S}}[F(\mathcal{A}(\mathcal{S}))]=\mathbb{E}_{\mathcal{S}, \mathcal{S}'}[f(\mathcal{A}(\mathcal{S}^{(i)}); \mathbf{z}_{i})]=\frac{1}{n}\sum_{i=1}^{n}\mathbb{E}_{\mathcal{S}, \mathcal{S}'}[f(\mathcal{A}(\mathcal{S}^{(i)}); \mathbf{z}_{i})].
\end{equation}
By leveraging facts \eqref{fact:1} and \eqref{fact:2}, we can equivalently reformulate the definition of average stability (Definition \ref{def:1}) as
\begin{equation}
	\label{stab_new}
	\left|\frac{1}{n}\sum_{i=1}^{n}\mathbb{E}_{\mathcal{S}, \mathcal{S}', \mathcal{A}}\left[f(\mathcal{A}(\mathcal{S}^{(i)});\mathbf{z}_{i})-f(\mathcal{A}(\mathcal{S});\mathbf{z}_{i})\right]\right|\leq \epsilon_{\text{stab}}.
\end{equation}
Furthermore, based on established facts \eqref{fact:1} and \eqref{fact:2}, it follows that
\begin{equation}
	\nonumber
	\begin{split}
		\epsilon_{\text{gen}}=\mathbb{E}_{\mathcal{S}, \mathcal{A}}\left[F(\mathcal{A}(\mathcal{S}))-F_{\mathcal{S}}(\mathcal{A}(\mathcal{S}))\right]&=\mathbb{E}_{\mathcal{S}, \mathcal{A}}\left[\mathbb{E}_{\mathcal{S}'}[f(\mathcal{A}(\mathcal{S}^{(i)}); \mathbf{z}_{i})]-\frac{1}{n}\sum_{i=1}^{n}f(\mathcal{A}(\mathcal{S}); \mathbf{z}_{i})\right]\\
		&=\frac{1}{n}\sum_{i=1}^{n}\mathbb{E}_{\mathcal{S}, \mathcal{S}', \mathcal{A}}\left[f(\mathcal{A}(\mathcal{S}^{(i)});\mathbf{z}_{i})-f(\mathcal{A}(\mathcal{S});\mathbf{z}_{i})\right]\\
		&\leq \epsilon_{\text{stab}}.
	\end{split}
\end{equation}
This implies that the new definition \eqref{stab_new} of average stability also has the property of Lemma \ref{lem:1}. 

$\hfill\blacksquare$ 

\section{Proof of Proposition \ref{thm:stab}}
According to Lemma \ref{lem:pi}, we have the following property in the non-expectation sense. That is, there exists $p_{i}, q_{i}\geq0$, such that
\begin{equation}
	\label{pq_supp}
	\|[x^{i}]\bm{\pi}(x)\|\leq p_{i},\quad \big\|\sqrt{\mathbf{A}}[x^{i}]\bm{\pi}(x)\big\| \leq q_{i},\quad i\in\{0, 1, \ldots, t\}.
\end{equation}
To prove Proposition \ref{thm:stab}, we need to analyze the terms $\mathbb{E}\|\mathbf{b}\|\|\mathbf{e}_{t}\|$ and $\mathbb{E}\|\sqrt{\mathbf{A}}\mathbf{e}_{t}\|\|\sqrt{\mathbf{A}}\mathbf{s}_{t}\|$ in the average stability \eqref{avg_stab}.
\begin{equation}
	\nonumber
	\begin{split}
		\mathbb{E}\|\mathbf{b}\|\|\mathbf{e}_{t}\|&=\mathbb{E}\|\mathbf{b}\|\|[x^{t}]\big(\bm{\phi}(x)\big)\|=\frac{\eta}{n}\mathbb{E}\|\mathbf{b}\|\|[x^{t}]\big(\bm{\pi}(x)\cdot\mathbf{u}(x)\big)\|\\
		&\overset{(i)}{=}\frac{\eta}{n}\mathbb{E}\big\|\mathbf{b}\big\|\Big\|\sum_{i=0}^{t}\left([x^{i}]\bm{\pi}(x)\right) \left([x^{t-i}]\mathbf{u}(x)\right)\Big\|\overset{(ii)}{=}\frac{\eta}{n}\mathbb{E}\big\|\mathbf{b}\big\|\Big\|\sum_{i=0}^{t-\tau-1}\left([x^{i}]\bm{\pi}(x)\right) (\bm{\xi}_{t-i-1}'-\bm{\xi}_{t-i-1})\Big\|\\
		&\overset{(iii)}{\leq}\frac{\eta}{n}\sum_{i=0}^{t-\tau-1}\mathbb{E}\big\|\mathbf{b}\big\|\Big\|\left([x^{i}]\bm{\pi}(x)\right) (\bm{\xi}_{t-i-1}'-\bm{\xi}_{t-i-1})\Big\|\overset{(iv)}{\leq}\frac{\eta}{n}\sum_{i=0}^{t-\tau-1}p_{i}\mathbb{E}\big\|\mathbf{b}\big\|\big\|\bm{\xi}_{t-i-1}'-\bm{\xi}_{t-i-1}\big\|\\
		&\overset{(v)}{\leq}\frac{2\eta r\sigma}{n}\sum_{i=0}^{t-\tau-1}p_{i},
	\end{split}
\end{equation}
where $(i)$ employs the Cauchy product for formal power series, i.e., $(\sum_{t}b_{t}x^{t})(\sum_{t}d_{t}x^{t})=\sum_{t}(\sum_{i=0}^{t}b_{i}d_{t-i})x^{t}$. $(ii)$ is based on the formulation of $\mathbf{u}(x)$, i.e., $\mathbf{u}(x)=\sum_{t=\tau}^{\infty}(\bm{\xi}_{t}'-\bm{\xi}_{t})x^{t+1}$. Here it should be noted that the subscripts of $\bm{\xi}$ start from $\tau$, so the upper summation bound is $t-\tau-1$ here. $(iii)$ utilizes the triangle inequality. $(iv)$ uses the Cauchy-Schwartz inequality and \eqref{pq_supp}. $(v)$ is based on Assumptions \ref{amp:2}, \ref{amp:3}. Analogously, we have 
\begin{equation}
	\nonumber
	\begin{split}
		&\mathbb{E}\big\|\sqrt{\mathbf{A}}\mathbf{e}_{t}\big\|\big\|\sqrt{\mathbf{A}}\mathbf{s}_{t}\big\|=\mathbb{E}\big\|\sqrt{\mathbf{A}}[x^{t}]\big(\bm{\phi}(x)\big)\big\|\big\|\sqrt{\mathbf{A}}[x^{t}]\big(\bm{\psi}(x)\big)\big\|=\frac{\eta}{n}\mathbb{E}\big\|\sqrt{\mathbf{A}}[x^{t}]\big(\bm{\pi}(x)\cdot\mathbf{u}(x)\big)\big\|\big\|\sqrt{\mathbf{A}}[x^{t}]\big(\bm{\pi}(x)\cdot(\mathbf{s}_0-\eta \mathbf{v}(x))\big)\big\|\\
		&\overset{(i)}{\leq}\frac{\eta}{n}\mathbb{E}\big\|\sqrt{\mathbf{A}}[x^{t}]\big(\bm{\pi}(x)\cdot\mathbf{u}(x)\big)\big\|\Big(\big\|\sqrt{\mathbf{A}}[x^{t}]\bm{\pi}(x)\mathbf{s}_0\big\|+\eta\big\|\sqrt{\mathbf{A}}[x^{t}]\big(\bm{\pi}(x)\cdot\mathbf{v}(x)\big)\big\|\Big)\\
		&\overset{(ii)}{\leq}\frac{\eta}{n}\mathbb{E}\Big\|\sum_{i=0}^{t-\tau-1}\left(\sqrt{\mathbf{A}}[x^{i}]\bm{\pi}(x)\right)(\bm{\xi}_{t-i-1}'-\bm{\xi}_{t-i-1})\Big\|\Big(\big\|\sqrt{\mathbf{A}}[x^{t}]\bm{\pi}(x)\mathbf{s}_0\big\|+\eta\Big\|\sum_{i=0}^{t-\tau-1}\left(\sqrt{\mathbf{A}}[x^{i}]\bm{\pi}(x)\right)(2\mathbf{b}+\bm{\xi}_{t-i-1}'+\bm{\xi}_{t-i-1})\Big\|\Big)\\
		&\overset{(iii)}{\leq}\frac{\eta}{n}\mathbb{E}\sum_{i=0}^{t-\tau-1}q_{i}\big\|\bm{\xi}_{t-i-1}'-\bm{\xi}_{t-i-1}\big\|\cdot\Big(q_{t}\big\|\mathbf{s}_0\big\|+\eta \sum_{i=0}^{t-\tau-1}q_{i}\big\|2\mathbf{b}+\bm{\xi}_{t-i-1}'+\bm{\xi}_{t-i-1}\big\|\Big)\\
		&\overset{(iv)}{\leq}\frac{2\eta q_{t}\|\mathbf{w}_0\|}{n}\sum_{i=0}^{t-\tau-1}q_{i}\mathbb{E}\big\|\bm{\xi}_{t-i-1}'-\bm{\xi}_{t-i-1}\big\|+\frac{\eta^{2}}{n} \mathbb{E}\sum_{i=0}^{t-\tau-1}q_{i}\big\|\bm{\xi}_{t-i-1}'-\bm{\xi}_{t-i-1}\big\|\cdot\sum_{i=0}^{t-\tau-1}q_{i}\big\|2\mathbf{b}+\bm{\xi}_{t-i-1}'+\bm{\xi}_{t-i-1}\big\|\Big)\\
		&\overset{(v)}{\leq}\frac{4\eta\sigma\|\mathbf{w}_0\|}{n}q_{t}\sum_{i=0}^{t-\tau-1}q_{i}+\frac{4\eta^{2}\sigma(r+\sigma)}{n} \Big(\sum_{i=0}^{t-\tau-1}q_{i}\Big)^{2},
	\end{split}
\end{equation}
where $(i)$ utilizes the linearity of the extraction operation $[x^{t}]$ and the triangle inequality. $(ii)$ follows from the Cauchy product for formal power series and the formulation of $\mathbf{u}(x), \mathbf{v}(x)$. $(iii)$ utilizes the triangle, Cauchy-Schwartz inequalities and \eqref{pq_supp}. $(iv)$ is based on the initialization \eqref{init}, i.e., $\mathbf{s}_0=\mathbf{w}_0+\mathbf{w}'_0$ and $\mathbf{w}_0=\mathbf{w}'_0$. $(v)$ uses the triangle inequality and Assumptions \ref{amp:2}, \ref{amp:3}. Substituting the aforementioned two terms into inequality \eqref{avg_stab} yields Proposition \ref{thm:stab}.
\begin{equation}
	\label{app:gen}
	\begin{split}
		\mathbb{E}\left|\frac{1}{n}\sum_{i=1}^{n} \left[f(\mathbf{w}_t';\mathbf{z}_{i})-f(\mathbf{w}_t;\mathbf{z}_{i})\right]\right|&\leq \frac{1}{2}\mathbb{E}\left\|\sqrt{\mathbf{A}}\mathbf{e}_{t}\right\|\left\|\sqrt{\mathbf{A}}\mathbf{s}_{t}\right\|+\mathbb{E}\|\mathbf{b}\|\|\mathbf{e}_{t}\|\\
		&\leq \frac{2\eta\sigma\|\mathbf{w}_0\|}{n}q_{t}\sum_{i=0}^{t-\tau-1}q_{i}+\frac{2\eta^{2}\sigma(r+\sigma)}{n} \Big(\sum_{i=0}^{t-\tau-1}q_{i}\Big)^{2}+\frac{2\eta r\sigma}{n}\sum_{i=0}^{t-\tau-1}p_{i}.
	\end{split}
\end{equation}


$\hfill\blacksquare$ 

\begin{remark}
	The detailed derivation of all formulas in Chapter \ref{sec:stab} is similar to the analysis of random delays in Appendix \ref{app:random}. For example, the study of generating functions \eqref{rec:phi} and \eqref{rec:psi} in the text can be referred to \eqref{app:phi} and \eqref{app:psi}, with the difference that \eqref{app:phi} and \eqref{app:psi} investigate the generating functions of the sequences with random delays.
\end{remark}

\section{Proof of Lemma \ref{lem:pi}}
The proof is derived from \cite[Lemma 1]{arjevani2020tight}. For the polynomial $\kappa(x)=1-x+\alpha x^{\tau+1}$, it has $\tau+1$ non-zero roots $x_1, \ldots, x_{\tau+1}$, and
\begin{equation}
	\nonumber
	\frac{1}{\kappa(x)}=\sum_{i=1}^{\tau+1}\frac{1}{\kappa^{\prime}(x_i)(x-x_i)}=\sum_{i=1}^{\tau+1}\frac{-1}{\kappa^{\prime}(x_i)x_i}\sum_{t=0}^{\infty}\left(\frac{x}{x_i}\right)^{t}, \text{~and~} \left[x^t\right]\left(\frac{1}{\kappa(x)}\right)=\sum_{i=1}^{\tau+1} \frac{-1}{\kappa^{\prime}\left(x_i\right) x_i^{t+1}}.
\end{equation}
Based on some standard tools from complex analysis, we have the following proposition
\begin{proposition}
	Let $\alpha \in (0, 1/20(\tau+1)]$, and assume $|x_1|\leq|x_2|\leq\ldots\leq|x_{\tau+1}|$, then
	\begin{enumerate}
		\item $x_1$ is a real scalar satisfying $1/x_1\leq1-\alpha$
		\item for $i>1$, $|1/x_{i}|\leq 1-\frac{3}{2(\tau+1)}$
		\item for $i\in [\tau+1]$, $|\kappa^{\prime}(x_{i})|>\frac{1}{2}$
	\end{enumerate}
\end{proposition}
then we have
\begin{equation}
	\nonumber
	\begin{split}
		\left|\left[x^t\right]\left(\frac{1}{\kappa(x)}\right)\right| \leq 2(1-\alpha)^{t+1}+2 \tau\left[1-\frac{3}{2(\tau+1)}\right]^{t+1} \leq 2(1-\alpha)^{t+1}\left[1+\tau \exp \left(-\frac{t+1}{\tau+1}\right)\right].
	\end{split}
\end{equation}
Furthermore, with $t_0=(\tau+1) \ln (2(\tau+1))$, we can derive
\begin{equation}
	\nonumber
	\left\{
	\begin{array}{ll}\vspace{1ex}
		\left|[x^t]\left(\frac{1}{\kappa(x)}\right)\right| \leq 1 & 0 \leq t \leq t_0-1, \\
		{\left|[x^t]\left(\frac{1}{\kappa(x)}\right)\right| \leq 3(1-\alpha)^{t+1}} & t \geq t_0.
	\end{array}\right.
\end{equation}
Extending $\kappa(x)=1-x+\alpha x^{\tau+1}$ to the matrix form $\bm{\pi}(x)=(\mathbf{I}-\mathbf{I}x+\eta \mathbf{A} x^{\tau+1})^{-1}$, then if $\forall j\in[d]$, let $\eta a_j \in (0, 1/20(\tau+1)]$, we have
\begin{equation}
	\label{app:pi}
	\left\{
	\begin{array}{ll}\vspace{1.3ex}
		\left\|[x^{t}]\bm{\pi}(x)\right\|  \leq 1   & \quad 0 \leq t \leq t_0-1, \\ 
		\left\|[x^{t}]\bm{\pi}(x)\right\| \leq 3\max_{j\in[d]}(1-\eta a_{j})^{t+1} & \quad t \geq t_0,
	\end{array}\right.
\end{equation}
where $a_{j}$ is the $j$-th eigenvalue of the positive semi-definite matrix $\mathbf{A}$ and satisfies that $0<a_{j}<\mu$. Let the learning rate satisfies $\eta \in (0, 1/20(\mu\tau+1)]$, then Lemma \ref{lem:pi} holds. Moreover, following \eqref{app:pi}, if $\eta \in (0, 1/20\mu(\tau+1)]$, we also have
\begin{equation}
	\label{app:pi-a}
	\left\{
	\begin{array}{ll}\vspace{1.5ex}
		\left\|\sqrt{\mathbf{A}}[x^{t}]\bm{\pi}(x)\right\| \leq \max\limits_{j\in[d]}\sqrt{a_{j}}   & 0 \leq t \leq t_0-1, \\ 
		\left\|\sqrt{\mathbf{A}}[x^{t}]\bm{\pi}(x)\right\| \leq 3\max\limits_{j\in[d]}\sqrt{a_{j}} (1-\eta a_{j})^{t+1} & t \geq t_0.
	\end{array}\right.
\end{equation}
More detailed derivations of this Lemma are available in \cite{arjevani2020tight}.

$\hfill\blacksquare$

\section{Proof of Theorem \ref{thm:1} and \ref{thm:2}}
\label{app:a4}
According to the average stability of delayed SGD (Proposition \ref{thm:stab}) and Lemma \ref{lem:1}, we first need to determine the constants $p_{i}, q_{i}$ in \eqref{pq}, and then bound the following three terms
\begin{equation}
	\nonumber
	\sum_{i=0}^{t-\tau-1}\left\|[x^{i}]\bm{\pi}(x)\right\|\leq\sum_{i=0}^{t-\tau-1}p_{i}, \quad \left\|\sqrt{\mathbf{A}}[x^{t}]\bm{\pi}(x)\right\|\leq q_{t}, \quad \text{and} \quad \sum_{i=0}^{t-\tau-1}\left\|\sqrt{\mathbf{A}}[x^{i}]\bm{\pi}(x)\right\|\leq\sum_{i=0}^{t-\tau-1}q_{i}.
\end{equation}
For these terms, we have to consider both convex and strongly convex cases separately. The difference is that when the quadratic function is strongly convex, the eigenvalues of the corresponding semi-positive definite matrix $\mathbf{A}$ further satisfy $\lambda\leq a_{j}\leq \mu, \forall j\in[d]$.

\subsection{Convex case}
Based on Lemma \ref{lem:pi}, the first term satisfies
\begin{equation}
	\label{c1}
	\sum_{i=0}^{t-\tau-1}\left\|[x^{i}]\bm{\pi}(x)\right\|\leq t-\tau.
\end{equation}
Following \eqref{app:pi-a}, the second and the third terms can be bounded as follows.
\begin{equation}
	\label{c2}
	\left\|\sqrt{\mathbf{A}}[x^{t}]\bm{\pi}(x)\right\|\leq 3\max\limits_{j\in[d]}\sqrt{a_j} (1-\eta a_j)^{t+1}\leq 3\max\limits_{j\in[d]} \sqrt{a_j} e^{-\eta a_j(t+1)}\leq\frac{3}{\sqrt{2\eta e(t+1)}}.
\end{equation}
The second inequality uses $(1+a)^t\leq e^{at}$. Note that the only stationary point of $\sqrt{a} e^{-\eta a(t+1)}$ is $a^{*}=\frac{1}{2\eta(t+1)}$, then we can obtain the third inequality. Similarly,
\begin{equation}
	\label{c3}
	\begin{split}
		\sum_{i=0}^{t-\tau-1}\left\|\sqrt{\mathbf{A}}[x^{i}]\bm{\pi}(x)\right\|&\leq\sum_{i=0}^{\lceil t_0\rceil-1}\left\|\sqrt{\mathbf{A}}[x^{i}]\bm{\pi}(x)\right\|+\sum_{i=\lceil t_0\rceil}^{t-\tau-1}\left\|\sqrt{\mathbf{A}}[x^{i}]\bm{\pi}(x)\right\|\\
		&\leq t_0\sqrt{\mu}+\sum_{i=\lceil t_0\rceil}^{t-\tau-1}\frac{3}{\sqrt{2e\eta (i+1)}}\\
		&\leq t_0\sqrt{\mu}+\frac{3}{\sqrt{2e\eta}}\sum_{i=\lceil t_0\rceil}^{t-\tau-1}\frac{1}{\sqrt{i+1}}\\
		&\leq t_0\sqrt{\mu}+6\sqrt{\frac{t-\tau}{2e\eta}}.
	\end{split}
\end{equation}
The last inequality is due to that
\begin{equation}
	\nonumber
	\sum_{i=\lceil t_0\rceil}^{t-\tau-1}\frac{1}{\sqrt{i+1}}\leq\sum_{i=\lceil t_0\rceil}^{t-\tau-1}\int_{i}^{i+1}\frac{1}{\sqrt{x}}dx\leq\sum_{i=\lceil t_0\rceil}^{t-\tau-1}2\sqrt{x} \Big{|}_{i}^{i+1}\leq 2\sqrt{x} \Big{|}_{t_0}^{t-\tau}\leq 2\sqrt{t-\tau}.
\end{equation}

\subsection{Strongly convex case}
In the strongly convex case, the eigenvalues of the semi-positive definite matrix $\mathbf{A}$ satisfy $\lambda\leq a_{j}\leq \mu, \forall j\in[d]$. Based on \eqref{app:pi} and \eqref{app:pi-a}, the first and second terms satisfy
\begin{equation}
	\label{sc1}
	\begin{split}
		\sum_{i=0}^{t-\tau-1}\left\|[x^{i}]\bm{\pi}(x)\right\|&\leq\sum_{i=0}^{\lceil t_0\rceil-1}\left\|[x^{i}]\bm{\pi}(x)\right\|+\sum_{i=\lceil t_0\rceil}^{t-\tau-1}\left\|[x^{i}]\bm{\pi}(x)\right\|\\
		&\leq t_0+3\sum_{i=\lceil t_0\rceil}^{t-\tau-1}\max_{j\in[d]}(1-\eta a_{j})^{t+1}\\
		&\leq t_0+3\sum_{i=\lceil t_0\rceil}^{t-\tau-1}(1-\eta\lambda)^{t+1}\leq t_0+\frac{3(1-\eta\lambda)^{t_0}}{\eta\lambda}\\
		&\leq t_0+\frac{3}{\eta\lambda},
	\end{split}
\end{equation}
and
\begin{equation}
	\label{sc2}
	\begin{split}
		\left\|\sqrt{\mathbf{A}}[x^{t}]\bm{\pi}(x)\right\|\leq 3\max\limits_{j\in[d]}\sqrt{a_j} (1-\eta a_j)^{t+1}\leq 3\sqrt{\mu} (1-\eta \lambda)^{t+1}\leq\frac{3\sqrt{\mu}}{e^{\eta \lambda(t+1)}}\leq 3\sqrt{\mu}.
	\end{split}
\end{equation}
For the third term, notice that for $i>t_0$,
\begin{equation}
	\nonumber
	\left\|\sqrt{\mathbf{A}}[x^{i}]\bm{\pi}(x)\right\|\leq 3\max\limits_{j\in[d]}\sqrt{a_j} (1-\eta a_j)^{i+1}.
\end{equation}
The only stationary point of $\sqrt{a} (1-\eta a)^{i+1}$ is $a^{*}=\frac{1}{2\eta(i+1)}$. It follows that if $\lambda\geq a^{*}$, or equivalently $i\geq\frac{1}{2\eta\lambda}-1$, then
\begin{equation}
	\nonumber
	\max_{\lambda\leq a \leq \mu}\sqrt{a} (1-\eta a)^{i+1}\leq\sqrt{\lambda} (1-\eta\lambda)^{i+1},
\end{equation}
and if $\lambda\leq a^{*}$, we have that
\begin{equation}
	\nonumber
	\max_{\lambda\leq a \leq \mu}\sqrt{a} (1-\eta a)^{i+1}\leq\sqrt{a^{*}} (1-\eta a^{*})^{i+1}\leq \sqrt{a^{*}} e^{-\eta (t+1)a^{*}}\leq\frac{1}{\sqrt{2e\eta (i+1)}}.
\end{equation}
Let $i_0=\frac{1}{2\eta\lambda}$ and if $i_0>t_0$, we can derive
\begin{equation}
	\nonumber
	\begin{split}
		\sum_{i=0}^{t-\tau-1}\left\|\sqrt{\mathbf{A}}[x^{i}]\bm{\pi}(x)\right\|&\leq\sum_{i=0}^{\lceil t_0\rceil-1}\left\|\sqrt{\mathbf{A}}[x^{i}]\bm{\pi}(x)\right\|+\sum_{i=\lceil t_0\rceil}^{\lceil i_0\rceil-1}\left\|\sqrt{\mathbf{A}}[x^{i}]\bm{\pi}(x)\right\|+\sum_{i=\lceil i_0\rceil}^{t-\tau-1}\left\|\sqrt{\mathbf{A}}[x^{i}]\bm{\pi}(x)\right\|\\
		&\leq t_0\sqrt{\mu}+\sum_{i=\lceil t_0\rceil}^{\lceil i_0\rceil-1}\frac{3}{\sqrt{2e\eta (i+1)}}+\sum_{i=\lceil i_0\rceil}^{t-\tau-1}3\sqrt{\lambda} (1-\eta\lambda)^{i+1}\\
		&\leq t_0\sqrt{\mu}+3\sqrt{\frac{2i_0}{e\eta}}+\frac{3(1-\eta\lambda)^{i_0}}{\eta\sqrt{\lambda}}\\
		&\leq t_0\sqrt{\mu}+\frac{6}{\eta\sqrt{e\lambda}},
	\end{split}
\end{equation}
where we use $(1-\eta\lambda)^{i_0}\leq e^{-\eta\lambda i_0}\leq 1/\sqrt{e}$ and
\begin{equation}
	\nonumber
	\sum_{i=\lceil t_0\rceil}^{\lceil i_0\rceil-1}\frac{1}{\sqrt{i+1}}\leq\sum_{i=\lceil t_0\rceil}^{\lceil i_0\rceil-1}\int_{i}^{i+1}\frac{1}{\sqrt{x}}dx\leq\sum_{i=\lceil t_0\rceil}^{\lceil i_0\rceil-1}2\sqrt{x} \Big{|}_{i}^{i+1}\leq 2\sqrt{x} \Big{|}_{t_0}^{i_0}\leq 2\sqrt{i_0}.
\end{equation}
For the case that $i_0<t_0$,
\begin{equation}
	\nonumber
	\begin{split}
		\sum_{i=0}^{t-\tau-1}\left\|\sqrt{\mathbf{A}}[x^{i}]\bm{\pi}(x)\right\|&\leq\sum_{i=0}^{\lceil t_0\rceil-1}\left\|\sqrt{\mathbf{A}}[x^{i}]\bm{\pi}(x)\right\|+\sum_{i=\lceil t_0\rceil}^{t-\tau-1}\left\|\sqrt{\mathbf{A}}[x^{i}]\bm{\pi}(x)\right\|\\
		&\leq t_0\sqrt{\mu}+\sum_{i=\lceil t_0\rceil}^{t-\tau-1}3\sqrt{\lambda} (1-\eta\lambda)^{i+1}\\
		&\leq t_0\sqrt{\mu}+\frac{3(1-\eta\lambda)^{t_0}}{\eta\sqrt{\lambda}}\leq t_0\sqrt{\mu}+\frac{3(1-\eta\lambda)^{i_0}}{\eta\sqrt{\lambda}}\\
		&\leq t_0\sqrt{\mu}+\frac{3}{\eta\sqrt{\lambda}}e^{-\eta\lambda i_0}\\
		&\leq t_0\sqrt{\mu}+\frac{3}{\eta\sqrt{e\lambda}}.
	\end{split}
\end{equation}
Then the third term in the strongly convex case is bounded as
\begin{equation}
	\label{sc3}
	\sum_{i=0}^{t-\tau-1}\left\|\sqrt{\mathbf{A}}[x^{i}]\bm{\pi}(x)\right\|\leq t_0\sqrt{\mu}+\frac{6}{\eta\sqrt{e\lambda}}.
\end{equation}
In the theorems, we require that $\eta\leq1/20\mu(\tau+1)$, $t\geq t_0=(\tau+1)\ln(2(\tau+1))$, then there is a fact that
\begin{equation}
	\label{app:fact}
	\eta t_{0}\leq \frac{\ln(2(\tau+1))}{20\mu}<\frac{\ln(\tau+1)}{10\mu}.
\end{equation}

\subsection{Generalization error in the convex case (Theorem \ref{thm:1})}
Based on Lemma \ref{lem:1} and Proposition \ref{thm:stab}, substituting items \eqref{c1}, \eqref{c2}, and \eqref{c3} into the inequality \eqref{app:gen}, the generalization error (or average stability) after $t$ iterations follows that
\begin{equation}
	\nonumber
	\begin{split}
		&\mathbb{E}_{\mathcal{S}, \mathcal{A}}[F(\mathbf{w}_{t})-F_{\mathcal{S}}(\mathbf{w}_{t})]\\
		\leq&\frac{2\eta r \sigma}{n}(t-\tau)+\frac{2\eta\sigma\left\|\mathbf{w}_0\right\|}{n}\frac{3}{\sqrt{2\eta e(t+1)}}\left(t_0\sqrt{\mu}+6\sqrt{\frac{t-\tau}{2e\eta}}\right)+\frac{2\eta^{2}\sigma(r+\sigma)}{n}\left(t_0\sqrt{\mu}+6\sqrt{\frac{t-\tau}{2e\eta}}\right)^{2}\\
		\leq&\frac{2\eta r\sigma}{n}(t-\tau)+\frac{6t_{0}\sigma\left\|\mathbf{w}_0\right\|}{n\sqrt{2e}}\sqrt{\frac{\mu\eta}{t+1}}+\frac{18\sigma\left\|\mathbf{w}_0\right\|}{ne}\sqrt{\frac{t-\tau}{t+1}}+\frac{2\eta^{2}t_{0}^{2}\mu\sigma(r+\sigma)}{n}+\frac{36\eta\sigma(r+\sigma)(t-\tau)}{ne}\\
		&+\frac{24\eta^{2}t_{0}\sigma(r+\sigma)}{n}\sqrt{\frac{\mu(t-\tau)}{2e\eta}}\\
		\overset{(\star)}{\leq}&\frac{r\sigma}{10n\mu\tau}(t-\tau)+\frac{6\sigma\left\|\mathbf{w}_0\right\| }{n\sqrt{2e}}\sqrt{\frac{(\tau+1)\ln^{2}(2(\tau+1))}{20(t+1)}}+\frac{18\sigma\left\|\mathbf{w}_0\right\|}{ne}+\frac{\sigma(r+\sigma)\ln^{2}(\tau+1)}{50n\mu}\\
		&+\frac{9\sigma(r+\sigma)}{5ne\mu\tau}(t-\tau)+\frac{12\sigma(r+\sigma)\ln(\tau+1)}{5n\mu}\sqrt{\frac{t-\tau}{40e\tau}}\\
		\leq&\frac{\sigma(r+\sigma)}{n\mu\tau}(t-\tau)+\frac{\sigma(r+\sigma)}{n\mu}\sqrt{t-\tau}+\frac{24\sigma\left\|\mathbf{w}_0\right\|}{ne}+\frac{\sigma(r+\sigma)\ln^{2}(\tau+1)}{n\mu}\\
		\leq&\frac{\sigma(r+\sigma)}{n\mu\tau}(t-\tau)+\frac{\sigma(r+\sigma)}{n\mu}\left[\sqrt{t-\tau}+12\mu\left\|\mathbf{w}_0\right\|+\ln^{2}(\tau+1)\right]\\
		\leq&\widetilde{\mathcal{O}}\left(\frac{t-\tau}{n\tau}\right),
	\end{split}
\end{equation}
where $(\star)$ uses the fact \eqref{app:fact} and the last inequality hides the log factors of $\tau$.

\subsection{Generalization error in the strongly convex case (Theorem \ref{thm:2})}
Similarly, by substituting the strongly convex items \eqref{sc1}, \eqref{sc2}, and \eqref{sc3} into the inequality \eqref{app:gen}, we can conclude that
\begin{equation}
	\nonumber
	\begin{split}
		&\mathbb{E}_{\mathcal{S}, \mathcal{A}}[F(\mathbf{w}_{t})-F_{\mathcal{S}}(\mathbf{w}_{t})]\\
		\leq&\frac{2\eta r\sigma}{n}(t_0+\frac{3}{\eta\lambda})+\frac{2\eta\sigma\left\|\mathbf{w}_0\right\|}{n}3\sqrt{\mu}\left(t_0\sqrt{\mu}+\frac{6}{\eta\sqrt{e\lambda}}\right)+\frac{2\eta^{2}\sigma(r+\sigma)}{n}\left(t_0\sqrt{\mu}+\frac{6}{\eta\sqrt{e\lambda}}\right)^{2}\\
		\leq&\frac{2\eta t_0 r\sigma}{n}+\frac{6r\sigma}{n\lambda}+\frac{6\eta\mu\sigma t_0\left\|\mathbf{w}_0\right\|}{n}+\frac{36\sigma\left\|\mathbf{w}_0\right\|}{n}\sqrt{\frac{\mu}{e\lambda}}+\frac{2\eta^{2}t_{0}^{2}\mu\sigma(r+\sigma)}{n}+\frac{72\sigma(r+\sigma)}{ne\lambda}+\frac{24\eta t_{0}\sigma(r+\sigma)}{n}\sqrt{\frac{\mu}{e\lambda}}\\
		\leq&\eta t_{0}\left[\frac{2r\sigma}{n}+\frac{6\mu\sigma\left\|\mathbf{w}_0\right\|}{n}+\frac{2\mu\sigma(r+\sigma)}{n}\eta t_0+\frac{24\sigma(r+\sigma)}{n}\sqrt{\frac{\mu}{e\lambda}}\right]+\frac{6r\sigma}{n\lambda}+\frac{36\sigma\left\|\mathbf{w}_0\right\|}{n}\sqrt{\frac{\mu}{e\lambda}}+\frac{72\sigma(r+\sigma)}{ne\lambda}\\
		\leq&\frac{2\sigma(r+\sigma)\eta t_{0}}{n}\left[1+3\mu\left\|\mathbf{w}_0\right\|+\mu\eta t_0+12\sqrt{\frac{\mu}{e\lambda}}\right]+\frac{42\sigma(r+\sigma)}{n\lambda}+\frac{36\sigma\left\|\mathbf{w}_0\right\|}{n}\sqrt{\frac{\mu}{e\lambda}}\\
		\overset{(\star)}{\leq}&\frac{\sigma(r+\sigma)\ln(\tau+1)}{5n\mu}\left[1+3\mu\left\|\mathbf{w}_0\right\|+\ln(\tau+1)+12\sqrt{\frac{\mu}{e\lambda}}\right]+\frac{42\sigma(r+\sigma)}{n\lambda}+\frac{36\sigma\left\|\mathbf{w}_0\right\|}{n}\sqrt{\frac{\mu}{e\lambda}}\\
		\leq&\widetilde{\mathcal{O}}\left(\frac{1}{n}\right).
	\end{split}
\end{equation}

$\hfill\blacksquare$

\section{Extension to random delays}
\label{app:random}
The delayed SGD with random delay $\tau_{t}$ performed as
\begin{equation}
	\nonumber
	\mathbf{w}_{t+1}=\mathbf{w}_{t}-\eta \nabla f(\mathbf{w}_{t-\tau_t}; \mathbf{z}_{i_t})=\mathbf{w}_{t}-\eta\left(\nabla F_{\mathcal{S}}(\mathbf{w}_{t-\tau_t})+\bm{\xi}_{t}\right).
\end{equation}
The sequences $\{\mathbf{e}_{t}\}_t$ and $\{\mathbf{s}_{t}\}_t$ are defined as $\mathbf{e}_{t}=\mathbf{w}_{t}-\mathbf{w}_{t}'$ and $\mathbf{s}_{t}=\mathbf{w}_{t}+\mathbf{w}_{t}'$, respectively. Following the initialization $\mathbf{w}_{0}=\mathbf{w}_{1}=\cdots=\mathbf{w}_{\overline{\tau}}$ and note that the two iterations start from the same model, i.e., $\mathbf{w}_{0}=\mathbf{w}_{0}'$, then we have
\begin{equation}
	\label{app:init}
	\begin{split}
		\mathbf{e}_{0}=\mathbf{e}_{1}=\cdots=\mathbf{e}_{\overline{\tau}}=\mathbf{0} ~~ \text{and} ~~ \mathbf{s}_{0}=\mathbf{s}_{1}=\cdots=\mathbf{s}_{\overline{\tau}}.
	\end{split}
\end{equation}
With probability $1-1/n$. $\mathbf{w}$ and $\mathbf{w}'$ evaluate gradient on the same data point $\mathbf{z}_{i_t}$
\begin{equation}
	\nonumber
	\begin{split}
		\mathbb{E}[\mathbf{e}_{t+1}]&=\mathbb{E}[\mathbf{w}_{t+1}-\mathbf{w}_{t+1}']=\mathbb{E}[\mathbf{w}_{t}-\mathbf{w}_{t}']-\eta\mathbb{E}\left[\nabla f(\mathbf{w}_{t-\tau_t}; \mathbf{z}_{i_t} )-\nabla f(\mathbf{w}_{t-\tau_t}'; \mathbf{z}_{i_t})\right]\\
		&=\mathbb{E}[\mathbf{e}_{t}]-\eta\mathbb{E}\left[\mathbf{x}_{i_t}\mathbf{x}_{i_t}^{\top}(\mathbf{w}_{t-\tau_t}-\mathbf{w}_{t-\tau_t}')\right]=\mathbb{E}[\mathbf{e}_{t}]-\eta \mathbb{E}[\mathbf{A} \mathbf{e}_{t-\tau_t}].
	\end{split}
\end{equation}
With probability $1/n$, the algorithm encounters different data $\mathbf{z}_i$ and $\mathbf{z}_i'$
\begin{equation}
	\nonumber
	\begin{split}
		\mathbb{E}[\mathbf{e}_{t+1}]&=\mathbb{E}[\mathbf{w}_{t+1}-\mathbf{w}_{t+1}']=\mathbb{E}[\mathbf{w}_{t}-\mathbf{w}_{t}']-\eta\mathbb{E}\left[\mathbf{A}\mathbf{w}_{t-\tau_t}+\mathbf{b}+\bm{\xi}_{t}-(\mathbf{A}\mathbf{w}_{t-\tau_t}'+\mathbf{b}+\bm{\xi}_{t}')\right]\\
		&=\mathbb{E}[\mathbf{e}_{t}]-\eta \mathbb{E}[\mathbf{A} \mathbf{e}_{t-\tau_t}]-\eta\mathbb{E}[\bm{\xi}_{t}-\bm{\xi}_{t}'].
	\end{split}
\end{equation}
Here, $\bm{\xi}_{t}'$ is the gradient noise with respect to the $F_{\mathcal{S}}(\mathbf{w}_{t-\tau_t}')$. Combining the two cases and taking the expectation of $\mathbf{e}_{t+1}$ with respect to the randomness of the algorithm, we can derive
\begin{equation}
	\label{app:et}
	\begin{split}
		\mathbb{E}[\mathbf{e}_{t+1}]&=\mathbb{E}[\mathbf{e}_{t}]-\eta \mathbb{E}[\mathbf{A} \mathbf{e}_{t-\tau_t}]-\frac{\eta}{n}\mathbb{E}[\bm{\xi}_{t}-\bm{\xi}_{t}']\\
		&=\mathbb{E}[\mathbf{e}_{t}]-\eta \mathbb{E}[\mathbf{A} \mathbf{e}_{t-\overline{\tau}}]-\eta\mathbb{E}[\frac{1}{n}(\bm{\xi}_{t}-\bm{\xi}_{t}')+\bm{\zeta}_{t}-\bm{\zeta}_{t}'],
	\end{split}
\end{equation}
where $\bm{\zeta}_{t}=\mathbf{A}(\mathbf{w}_{t-\tau_t}-\mathbf{w}_{t-\overline{\tau}})$ and $\bm{\zeta}_{t}'=\mathbf{A}(\mathbf{w}_{t-\tau_t}'-\mathbf{w}_{t-\overline{\tau}}')$. Due to the convergence property of delayed SGD and the bounded delay Assumption \ref{amp:4}, it is reasonable to conclude that $\bm{\zeta}_{t}$ and $\bm{\zeta}_{t}'$ are bounded, i.e., $\mathbb{E}_{\mathcal{S}, \mathcal{A}}[\|\bm{\zeta}_{t}\|], \mathbb{E}_{\mathcal{S}, \mathcal{A}}[\|\bm{\zeta}_{t}'\|]\leq \rho$, where $\rho\in\mathbb{R}$ is constant. And since the datasets $\mathcal{S}$ and $\mathcal{S}^{(i)}$ differ by only one data sample, the difference $\bm{\zeta}_{t}-\bm{\zeta}_{t}'$ is roughly of order $\mathcal{O}(1/n)$. Then we have that $\mathbb{E}_{\mathcal{S}, \mathcal{A}}[\|\bm{\zeta}_{t}-\bm{\zeta}_{t}'\|]\leq 2\rho/n$. For the sequence $\{\mathbf{s}_{t}\}_{t}$, we do not need to examine them separately.
\begin{equation}
	\label{app:st}
	\begin{split}
		\mathbb{E}[\mathbf{s}_{t+1}]&=\mathbb{E}[\mathbf{w}_{t+1}+\mathbf{w}_{t+1}']=\mathbb{E}[\mathbf{w}_{t}+\mathbf{w}_{t}']-\eta\mathbb{E}\left[\mathbf{A}\mathbf{w}_{t-\tau_t}+\mathbf{b}+\bm{\xi}_{t}+(\mathbf{A}\mathbf{w}_{t-\tau_t}'+\mathbf{b}+\bm{\xi}_{t}')\right]\\
		&=\mathbb{E}[\mathbf{s}_{t}]-\eta \mathbb{E}[\mathbf{A} \mathbf{s}_{t-\tau_t}]-\eta\mathbb{E}[2\mathbf{b}+\bm{\xi}_{t}+\bm{\xi}_{t}']\\
		&=\mathbb{E}[\mathbf{s}_{t}]-\eta \mathbb{E}[\mathbf{A} \mathbf{s}_{t-\overline{\tau}}]-\eta\mathbb{E}[2\mathbf{b}+\bm{\xi}_{t}+\bm{\xi}_{t}'+\bm{\zeta}_{t}+\bm{\zeta}_{t}'].
	\end{split}
\end{equation}
Here we also have $\mathbb{E}_{\mathcal{S}, \mathcal{A}}\|\bm{\zeta}_{t}+\bm{\zeta}_{t}'\|\leq 2\rho$ and let
\begin{equation}
	\nonumber
	\bm{\phi}(x)=\sum_{t=0}^\infty \mathbf{e}_t x^t \quad \quad \bm{\psi}(x)=\sum_{t=0}^\infty \mathbf{s}_t x^t.
\end{equation}
For the sake of brevity, we omit the expectation notation in the derivation of the generating function. Based on the recurrence formula \eqref{app:et},
\begin{equation}
	\label{app:phi}
	\begin{split}
		\bm{\phi}(x)&=\sum_{t=0}^{\overline{\tau}} \mathbf{e}_t x^t+\sum_{t=\overline{\tau}+1}^{\infty}\mathbf{e}_{t} x^t\\
		&\overset{(\star)}{=}\sum_{t=\overline{\tau}+1}^{\infty}\left[\mathbf{e}_{t-1}-\eta \mathbf{A} \mathbf{e}_{t-\overline{\tau}-1}-\frac{\eta}{n}(\bm{\xi}_{t-1}-\bm{\xi}_{t-1}')-\eta(\bm{\zeta}_{t-1}-\bm{\zeta}_{t-1}')\right] x^t\\
		&=\sum_{t=\overline{\tau}+1}^{\infty} \mathbf{e}_{t-1} x^t-\eta \mathbf{A} \sum_{t=\overline{\tau}+1}^{\infty} \mathbf{e}_{t-\overline{\tau}-1} x^t -\eta\sum_{t=\overline{\tau}+1}^{\infty}\Big[\frac{1}{n}(\bm{\xi}_{t-1}-\bm{\xi}_{t-1}')+\bm{\zeta}_{t-1}-\bm{\zeta}_{t-1}'\Big]x^{t}\\
		&=x\sum_{t=\overline{\tau}}^{\infty} \mathbf{e}_{t} x^t-\eta \mathbf{A}x^{\overline{\tau}+1} \sum_{t=0}^{\infty} \mathbf{e}_{t} x^t-\eta\sum_{t=\overline{\tau}+1}^{\infty}\Big[\frac{1}{n}(\bm{\xi}_{t-1}-\bm{\xi}_{t-1}')+\bm{\zeta}_{t-1}-\bm{\zeta}_{t-1}'\Big]x^{t}\\
		&\overset{(\star)}{=}x\sum_{t=0}^{\infty} \mathbf{e}_{t} x^t-\eta \mathbf{A}x^{\overline{\tau}+1} \sum_{t=0}^{\infty} \mathbf{e}_{t} x^t-\eta\sum_{t=\overline{\tau}+1}^{\infty}\Big[\frac{1}{n}(\bm{\xi}_{t-1}-\bm{\xi}_{t-1}')+\bm{\zeta}_{t-1}-\bm{\zeta}_{t-1}'\Big]x^{t}\\
		&=x\bm{\phi}(x)-\eta \mathbf{A} x^{\overline{\tau}+1} \bm{\phi}(x)-\eta\sum_{t=\overline{\tau}}^{\infty}\Big[\frac{1}{n}(\bm{\xi}_{t}-\bm{\xi}_{t}')+\bm{\zeta}_{t}-\bm{\zeta}_{t}'\Big]x^{t+1},
	\end{split}
\end{equation}
where $(\star)$ uses the initialization \eqref{app:init}. Similarly, based on the recurrence relation \eqref{app:st}, we have
\begin{equation}
	\label{app:psi}
	\begin{split}
		\bm{\psi}(x)&=\sum_{t=0}^{\overline{\tau}} \mathbf{s}_t x^t+\sum_{t=\overline{\tau}+1}^{\infty}\mathbf{s}_{t} x^t\\
		&=\sum_{t=0}^{\overline{\tau}} \mathbf{s}_t x^t+\sum_{t=\overline{\tau}+1}^{\infty}\Big[\mathbf{s}_{t-1}-\eta \mathbf{A} \mathbf{s}_{t-\overline{\tau}-1}-\eta(2\mathbf{b}+\bm{\xi}_{t-1}+\bm{\xi}_{t-1}'+\bm{\zeta}_{t-1}+\bm{\zeta}_{t-1}')\Big] x^t\\
		&=\sum_{t=0}^{\overline{\tau}} \mathbf{s}_t x^t+\sum_{t=\overline{\tau}+1}^{\infty} \mathbf{s}_{t-1} x^t-\eta \mathbf{A} \sum_{t=\overline{\tau}+1}^{\infty} \mathbf{s}_{t-\overline{\tau}-1} x^{t} -\eta\sum_{t=\overline{\tau}}^{\infty}(2\mathbf{b}+\bm{\xi}_{t}+\bm{\xi}_{t}'+\bm{\zeta}_{t}+\bm{\zeta}_{t}')x^{t+1}\\
		&=\sum_{t=0}^{\overline{\tau}} \mathbf{s}_t x^t+x\sum_{t=\overline{\tau}}^{\infty} \mathbf{s}_{t} x^t-\eta \mathbf{A}x^{\overline{\tau}+1} \sum_{t=0}^{\infty} \mathbf{s}_{t} x^t -\eta\sum_{t=\overline{\tau}}^{\infty}(2\mathbf{b}+\bm{\xi}_{t}+\bm{\xi}_{t}'+\bm{\zeta}_{t}+\bm{\zeta}_{t}')x^{t+1}\\
		&=\sum_{t=0}^{\overline{\tau}} \mathbf{s}_t x^t-x\sum_{t=0}^{\overline{\tau}-1} \mathbf{s}_{t} x^t+x\sum_{t=0}^{\infty} \mathbf{s}_{t} x^t-\eta \mathbf{A}x^{\overline{\tau}+1} \sum_{t=0}^{\infty} \mathbf{s}_{t} x^t-\eta\sum_{t=\overline{\tau}}^{\infty}(2\mathbf{b}+\bm{\xi}_{t}+\bm{\xi}_{t}'+\bm{\zeta}_{t}+\bm{\zeta}_{t}')x^{t+1}\\
		&\overset{(\star)}{=}\mathbf{s}_0+x\bm{\psi}(x)-\eta \mathbf{A} x^{\overline{\tau}+1} \bm{\psi}(x)-\eta \sum_{t=\overline{\tau}}^{\infty}(2\mathbf{b}+\bm{\xi}_{t}+\bm{\xi}_{t}'+\bm{\zeta}_{t}+\bm{\zeta}_{t}')x^{t+1}.
	\end{split}
\end{equation}
Abbreviating the last terms in equations \eqref{app:phi} and \eqref{app:psi} as the power series $\mathbf{u}(x)$ and $\mathbf{v}(x)$, i.e.
\begin{equation}
	\nonumber
	\mathbf{u}(x)=\sum_{t=\overline{\tau}}^{\infty}\Big[\frac{1}{n}(\bm{\xi}_{t}'-\bm{\xi}_{t})+\bm{\zeta}_{t}'-\bm{\zeta}_{t}\Big]x^{t+1}, \quad \mathbf{v}(x)=\sum_{t=\overline{\tau}}^{\infty}(2\mathbf{b}+\bm{\xi}_{t}+\bm{\xi}_{t}'+\bm{\zeta}_{t}+\bm{\zeta}_{t}')x^{t+1},
\end{equation}
and rearranging terms in equations \eqref{app:phi} and \eqref{app:psi} gives
\begin{equation}
	\nonumber
	(\mathbf{I}-\mathbf{I}x+\eta \mathbf{A} x^{\overline{\tau}+1})\bm{\phi}(x)=\eta\mathbf{u}(x) \quad \text{and} \quad (\mathbf{I}-\mathbf{I}x+\eta \mathbf{A} x^{\overline{\tau}+1})\bm{\psi}(x)=\mathbf{s}_0-\eta \mathbf{v}(x).
\end{equation}
Let $\bm{\pi}(x)=(\mathbf{I}-\mathbf{I}x+\eta \mathbf{A} x^{\overline{\tau}+1})^{-1}$, then the following concise properties are observed for the generating functions $\bm{\phi}(x)$ and $\bm{\psi}(x)$, i.e.,
\begin{equation}
	\nonumber
	\bm{\phi}(x)=\eta\bm{\pi}(x)\cdot\mathbf{u}(x) \quad \text{and} \quad \bm{\psi}(x)=\bm{\pi}(x)\cdot\left[\mathbf{s}_0-\eta \mathbf{v}(x)\right].
\end{equation}
To analyze average stability \eqref{avg_stab}, we need to extract the corresponding coefficients in generating functions, i.e., $\mathbf{e}_{t}=[x^{t}]\bm{\phi}(x)$, $\mathbf{s}_{t}=[x^{t}]\bm{\psi}(x)$. The specific derivations are as follows
\begin{equation}
	\nonumber
	\begin{split}
		\mathbb{E}\|\mathbf{b}\|\|\mathbf{e}_{t}\|&=\eta\mathbb{E}\|\mathbf{b}\|\left\|[x^{t}](\bm{\pi}(x) \mathbf{u}(x))\right\|\overset{(i)}{=}\eta\mathbb{E}\|\mathbf{b}\|\left\|\sum_{i=0}^{t-\overline{\tau}-1}\big([x^{i}]\bm{\pi}(x)\big)\Big[\frac{1}{n}(\bm{\xi}_{t-i-1}'-\bm{\xi}_{t-i-1})+\bm{\zeta}_{t-i-1}'-\bm{\zeta}_{t-i-1}\Big]\right\|\\
		&\overset{(ii)}{\leq}\frac{2\eta r(\sigma+\rho)}{n}\sum_{i=0}^{t-\overline{\tau}-1}p_{i},
	\end{split}
\end{equation}
where $(i)$ employs the Cauchy product for formal power series, i.e., $(\sum_{t}b_{t}x^{t})(\sum_{t}d_{t}x^{t})=\sum_{t}(\sum_{i=0}^{t}b_{i}d_{t-i})x^{t}$. $(ii)$ uses Assumptions \ref{amp:2}, \ref{amp:3}, \eqref{pq}, and $\mathbb{E}_{\mathcal{S}, \mathcal{A}}[\|\bm{\zeta}_{t}-\bm{\zeta}_{t}'\|]\leq 2\rho/n$. Similarly, we have
\begin{equation}
	\nonumber
	\begin{split}
		\mathbb{E}\|\sqrt{\mathbf{A}}\mathbf{e}_{t}\|\|\sqrt{\mathbf{A}}\mathbf{s}_{t}\|&=\mathbb{E}\eta\left\|\sqrt{\mathbf{A}}[x^{t}](\bm{\pi}(x) \mathbf{u}(x))\right\|\left\|\sqrt{\mathbf{A}}[x^{t}](\bm{\pi}(x)(\mathbf{s}_0-\eta \mathbf{v}(x)))\right\|\\
		&\overset{(i)}{\leq}\mathbb{E}\eta\left\|\sqrt{\mathbf{A}}[x^{t}](\bm{\pi}(x) \mathbf{u}(x))\right\|\left(\left\|\sqrt{\mathbf{A}}[x^{t}]\bm{\pi}(x)\mathbf{s}_0\right\|+\eta\left\|\sqrt{\mathbf{A}}[x^{t}](\bm{\pi}(x)\mathbf{v}(x))\right\|\right)\\
		&\overset{(ii)}{\leq}\mathbb{E}\eta\left\|\sum_{i=0}^{t-\overline{\tau}-1}\sqrt{\mathbf{A}}\left([x^{i}]\bm{\pi}(x)\right) \Big[\frac{1}{n}(\bm{\xi}_{t-i-1}'-\bm{\xi}_{t-i-1})+\bm{\zeta}_{t-i-1}'-\bm{\zeta}_{t-i-1}\Big]\right\|\cdot\Bigg(\left\|\sqrt{\mathbf{A}}[x^{t}]\bm{\pi}(x)\mathbf{s}_0\right\|\\
		&\qquad+\eta\left\|\sum_{i=0}^{t-\overline{\tau}-1}\sqrt{\mathbf{A}}[x^{i}]\bm{\pi}(x)\left(2\mathbf{b}+\bm{\xi}_{t-i-1}+\bm{\xi}_{t-i-1}'+\bm{\zeta}_{t-i-1}+\bm{\zeta}_{t-i-1}'\right)\right\|\Bigg)\\
		&\overset{(iii)}{\leq}\frac{2\eta(\sigma+\rho)}{n}\sum_{i=0}^{t-\overline{\tau}-1}q_{i}\left(q_{t}\left\|\mathbf{s}_0\right\|+2\eta(r+\sigma+\rho)\sum_{i=0}^{t-\overline{\tau}-1}q_{i}\right)\\
		&\leq\frac{2\eta(\sigma+\rho)\left\|\mathbf{s}_0\right\|}{n}q_{t}\sum_{i=0}^{t-\overline{\tau}-1}q_{i}+\frac{4\eta^{2}(\sigma+\rho)(r+\sigma+\rho)}{n}\left(\sum_{i=0}^{t-\overline{\tau}-1}q_{i}\right)^{2}.
	\end{split}
\end{equation}
where $(i)$ uses the linearity of the extraction operation $[x^{t}]$. $(ii)$ followed by the Cauchy product for formal power series. $(iii)$ is based on Assumptions \ref{amp:2}, \ref{amp:3}, \eqref{pq}, $\mathbb{E}_{\mathcal{S}, \mathcal{A}}[\|\bm{\zeta}_{t}\|]\leq \rho$, and $\mathbb{E}_{\mathcal{S}, \mathcal{A}}[\|\bm{\zeta}_{t}-\bm{\zeta}_{t}'\|]\leq 2\rho/n$. Finally, we derive the average stability bound of delayed SGD with random delays.
\begin{equation}
	\label{app:stab-random}
	\begin{split}
		\epsilon_{\text{stab}}\leq\frac{2\eta r(\sigma+\rho)}{n}\sum_{i=0}^{t-\overline{\tau}-1}p_{i}+\frac{\eta(\sigma+\rho)\left\|\mathbf{s}_0\right\|}{n}q_{t}\sum_{i=0}^{t-\overline{\tau}-1}q_{i}+\frac{2\eta^{2}(\sigma+\rho)(r+\sigma+\rho)}{n}\left(\sum_{i=0}^{t-\overline{\tau}-1}q_{i}\right)^{2}.
	\end{split}
\end{equation}
Compared to fixed delays, analyzing the generalization error bounds for SGD with random delays requires only minor modifications.
\begin{lemma}
	\label{lem:pi_random}
	For the power series $\bm{\pi}(x)=(\mathbf{I}-\mathbf{I}x+\eta \mathbf{A} x^{\overline{\tau}+1})^{-1}$, let $\eta \in (0, 1/\mu\overline{\tau}]$, then
	\begin{equation}
		\nonumber
		\|[x^{t}]\bm{\pi}(x)\|\leq 1, ~~ \forall t\geq0.
	\end{equation}
	Further, if $\eta \in (0, 1/20\mu(\overline{\tau}+1)]$, we have
	\begin{equation}
		\nonumber
		\left\{
		\begin{array}{ll}\vspace{1.5ex}
			\left\|[x^{t}]\bm{\pi}(x)\right\| \leq 1   & 0 \leq t \leq t_0-1, \\ 
			\left\|[x^{t}]\bm{\pi}(x)\right\| \leq 3\max_{j\in[d]}(1-\eta a_{j})^{t+1} & t \geq t_0.
		\end{array}\right.
	\end{equation}
	where $t_0=(\overline{\tau}+1)\ln(2(\overline{\tau}+1))$ and $a_{j}$ is the $j$-th eigenvalue of the positive semi-definite matrix $\mathbf{A}$. Note that this property also holds in the expected sense with respect to the data set $\mathcal{S}$.
\end{lemma}
Similar to section \ref{app:a4}, we can bound the corresponding three terms in the convex and strongly convex cases as follows.
\begin{equation}
	\nonumber
	\left\{
	\begin{array}{ll}\vspace{1.5ex}
		\sum\limits_{i=0}^{t-\overline{\tau}-1}\left\|[x^{i}]\bm{\pi}(x)\right\| \leq t-\overline{\tau} &\text{convex},\\
		\sum\limits_{i=0}^{t-\overline{\tau}-1}\left\|[x^{i}]\bm{\pi}(x)\right\| \leq t_0+\frac{3}{\eta\lambda}  &\lambda\text{-strongly convex}.
	\end{array}\right.
\end{equation}
\begin{equation}
	\nonumber
	\left\{
	\begin{array}{ll}\vspace{1.5ex}
		\left\|\sqrt{\mathbf{A}}[x^{t}]\bm{\pi}(x)\right\| \leq \frac{3}{\sqrt{2\eta e(t+1)}} &\text{convex},\\
		\left\|\sqrt{\mathbf{A}}[x^{t}]\bm{\pi}(x)\right\| \leq 3\sqrt{\mu}  &\lambda\text{-strongly convex}.
	\end{array}\right.
\end{equation}
\begin{equation}
	\nonumber
	\left\{
	\begin{array}{ll}\vspace{1.5ex}
		\sum\limits_{i=0}^{t-1-\overline{\tau}}\left\|\sqrt{\mathbf{A}}[x^{i}]\bm{\pi}(x)\right\| \leq t_0\sqrt{\mu}+6\sqrt{\frac{t-\overline{\tau}}{2e\eta}} &\text{convex},\\
		\sum\limits_{i=0}^{t-1-\overline{\tau}}\left\|\sqrt{\mathbf{A}}[x^{i}]\bm{\pi}(x)\right\| \leq t_0\sqrt{\mu}+\frac{6}{\eta\sqrt{e\lambda}}  &\lambda\text{-strongly convex}.
	\end{array}\right.
\end{equation}
According to the average stability \eqref{app:stab-random} and Lemma \ref{lem:1}, we can derive the generalization error of the delayed SGD algorithm with random delays. For the convex problem, we substitute the three items into the inequality \eqref{app:stab-random}, resulting in
\begin{equation}
	\nonumber
	\begin{split}
		\mathbb{E}_{\mathcal{S}, \mathcal{A}}[F(\mathbf{w}_{t})-F_{\mathcal{S}}(\mathbf{w}_{t})]&\leq\frac{2\eta r (\sigma+\rho)}{n}(t-\overline{\tau})+\frac{\eta(\sigma+\rho)\left\|\mathbf{s}_0\right\|}{n}\frac{3}{\sqrt{2\eta e(t+1)}}\left(t_0\sqrt{\mu}+6\sqrt{\frac{t-\overline{\tau}}{2e\eta}}\right)\\
		&\quad+\frac{2\eta^{2}(\sigma+\rho)(r+\sigma+\rho)}{n}\left(t_0\sqrt{\mu}+6\sqrt{\frac{t-\overline{\tau}}{2e\eta}}\right)^{2}\\
		&\leq\frac{(\sigma+\rho)(r+\sigma+\rho)}{n\mu\overline{\tau}}(t-\overline{\tau})+\frac{(\sigma+\rho)(r+\sigma+\rho)}{n\mu}\left[\sqrt{t-\overline{\tau}}+12\mu\left\|\mathbf{w}_0\right\|+\ln^{2}(\overline{\tau}+1)\right]\\
		&\leq\widetilde{\mathcal{O}}\left(\frac{t-\overline{\tau}}{n\overline{\tau}}\right).
	\end{split}
\end{equation}
Here we use $\eta\leq1/20\mu(\overline{\tau}+1)$, $t\geq t_0=(\overline{\tau}+1)\ln(2(\overline{\tau}+1))$, and the fact
\begin{equation}
	\label{app:fact_random}
	\eta t_{0}\leq \frac{\ln(2(\overline{\tau}+1))}{20\mu}<\frac{\ln(\overline{\tau}+1)}{10\mu}.
\end{equation}
If the quadratic loss function is strongly convex, we can obtain 
\begin{equation}
	\nonumber
	\begin{split}
		\mathbb{E}_{\mathcal{S}, \mathcal{A}}[F(\mathbf{w}_{t})-F_{\mathcal{S}}(\mathbf{w}_{t})]&\leq\frac{2\eta r(\sigma+\rho)}{n}(t_0+\frac{3}{\eta\lambda})+\frac{\eta(\sigma+\rho)\left\|\mathbf{s}_0\right\|}{n}3\sqrt{\mu}\left(t_0\sqrt{\mu}+\frac{6}{\eta\sqrt{e\lambda}}\right)\\
		&\quad+\frac{2\eta^{2}(\sigma+\rho)(r+\sigma+\rho)}{n}\left(t_0\sqrt{\mu}+\frac{6}{\eta\sqrt{e\lambda}}\right)^{2}\\
		&\leq\frac{(\sigma+\rho)(r+\sigma+\rho)\ln(\overline{\tau}+1)}{5n\mu}\left[1+3\mu\left\|\mathbf{w}_0\right\|+\ln(\overline{\tau}+1)+12\sqrt{\frac{\mu}{e\lambda}}\right]\\
		&\quad+\frac{42(\sigma+\rho)(r+\sigma+\rho)}{n\lambda}+\frac{36(\sigma+\rho)\left\|\mathbf{w}_0\right\|}{n}\sqrt{\frac{\mu}{e\lambda}}\\
		&\leq\widetilde{\mathcal{O}}\left(\frac{1}{n}\right).
	\end{split}
\end{equation}

$\hfill\blacksquare$

\end{document}